\documentclass[]{interact}% add a journal-specific option here only if the IFA asks (e.g. [largeformat])

\newif\ifAnonymous
\Anonymousfalse

\usepackage{natbib}
\bibpunct[, ]{(}{)}{;}{a}{}{,}

\usepackage{amsmath}
\usepackage{amssymb}
\usepackage{graphicx}
\usepackage{booktabs}
\usepackage{longtable}
\usepackage{multirow}
\usepackage{adjustbox}
\usepackage{changepage}
\usepackage{algorithm}
\usepackage{algorithmic}
\usepackage{rotating}
\usepackage{textcomp}
\usepackage[caption=false]{subfig}% Interact's preferred sub-figure package
\usepackage{url}
\usepackage{placeins}% provides \FloatBarrier (used in body.tex)

\usepackage{xcolor}

\providecommand{\extralength}{0pt}
\providecommand{\fulllength}{\textwidth}
\providecommand{\reftitle}[1]{}
\providecommand{\externalbibliography}[1]{}
\usepackage{hyperref}
\hypersetup{colorlinks=true, citecolor=blue, linkcolor=blue, urlcolor=blue}

\newif\ifReview
\Reviewfalse
\ifReview
\usepackage{lineno}
\newcommand*\patchAmsMathEnvironmentForLineno[1]{%
  \expandafter\let\csname old#1\expandafter\endcsname\csname #1\endcsname
  \expandafter\let\csname oldend#1\expandafter\endcsname\csname end#1\endcsname
  \renewenvironment{#1}%
    {\linenomath\csname old#1\endcsname}%
    {\csname oldend#1\endcsname\endlinenomath}}
\newcommand*\patchBothAmsMathEnvironmentsForLineno[1]{%
  \patchAmsMathEnvironmentForLineno{#1}%
  \patchAmsMathEnvironmentForLineno{#1*}}
\AtBeginDocument{%
  \patchBothAmsMathEnvironmentsForLineno{equation}%
  \patchBothAmsMathEnvironmentsForLineno{align}%
  \patchBothAmsMathEnvironmentsForLineno{flalign}%
  \patchBothAmsMathEnvironmentsForLineno{alignat}%
  \patchBothAmsMathEnvironmentsForLineno{gather}%
  \patchBothAmsMathEnvironmentsForLineno{multline}%
}
\fi

\begin{document}
\ifReview\linenumbers\fi

\articletype{Research Article}% adjust per the journal if needed

\title{iSAGE: A Human-in-the-Loop Framework for Remote Sensing Semantic Segmentation via Sparse Point Supervision}

\ifAnonymous
\author{\name{Anonymous}}
\else
\author{\name{Osmar Luiz Ferreira de Carvalho\textsuperscript{a}, Osmar Ab\'ilio de Carvalho J\'unior\textsuperscript{b}\thanks{CONTACT Osmar Ab\'ilio de Carvalho J\'unior. Email: osmarjr@unb.br}, Anesmar Olino de Albuquerque\textsuperscript{b} and Daniel Guerreiro e Silva\textsuperscript{a}}
\affil{\textsuperscript{a}Department of Electrical Engineering, University of Bras\'ilia, Bras\'ilia, DF, Brazil; \textsuperscript{b}Department of Geography, University of Bras\'ilia, Bras\'ilia, DF, Brazil}}
\fi

\maketitle

\begin{abstract}
Training semantic segmentation models relies on pixel-level labels, which is especially costly in remote sensing, since most problems require a new dataset as targets vary widely with resolution, sensors, and region. Approaches to reduce this cost, such as sparse annotations and human-in-the-loop methods, are increasingly common, but most studies add extra machinery that expands few labeled pixels into a denser training signal. This expansion relies on the model's own predictions, where a confident prediction looks the same whether it is correct or incorrect. The hypothesis here is that these confident errors are the most valuable pixels to label, and that a human examining the image can identify them directly. Testing it in isolation has never been done, because that requires a single integrated loop of annotation, training, and data management that no existing tool provides. We propose iSAGE (Iterative Sparse Annotation Guided by Expert), a novel open-source framework that closes this gap using the human alone, with no automatic label expansion. Each round runs three steps inside a single environment: (i) the annotator reviews the model's
  predictions over the image in an interactive annotation interface and clicks the confident errors, (ii) those clicks train the model under an error-weighted loss that emphasizes them, and (iii) the updated model surfaces the next errors for the following round. The experiments use a minimum-effort regime (at most one labeled pixel per class in each frame) on two datasets with complementary roles. BsB Aerial, built for this study, isolates four contrasting class types (cars, roads, buildings, and permeable areas) and recovers 97.2\% of dense performance (74.79\% mIoU from 0.040\% of the pixels), showing that amorphous regions (e.g., permeable areas) have high IoU from the first round and gain little across iterations (+9.7 IoU), whereas small objects (e.g., cars) improve substantially (+46.6 IoU over five rounds). ISPRS Vaihingen provides external validation on a public benchmark, where iSAGE reaches 76.78\% mIoU from 0.011\% of
  the pixels, matching the dense baseline (76.65\%) and surpassing published weakly-supervised methods by nearly 4 percentage points in mIoU. Four automatic selection strategies run through the same loop (oracle entropy, pseudo-labeling, CRF propagation, and random sampling) all plateau 7.4 to 14.5 points below iSAGE even at far higher budgets, showing the gap is methodological rather than a matter of budget. Across 31 surveyed methods, iSAGE is the only human-in-the-loop framework that adds no automatic label generation, supporting the claim that labeling confident errors directly is enough.
\end{abstract}

\begin{keywords}
remote sensing; semantic segmentation; sparse annotation; human-in-the-loop; active learning
\end{keywords}

% Body content (same body.tex as the ESWA/Elsevier version)
\clearpage  % start the Introduction at the top of a fresh page
\section{Introduction}\label{sec:intro}
How much data is needed to train computer vision models? Although the answer is task-, domain-, and target-dependent, the prevailing assumption is that extensive labeled data is a prerequisite. This belief has been reinforced by large-scale benchmark datasets such as Common Objects in Context (COCO) \citep{Lin2014Microsoft}, Cityscapes \citep{cordts2016cityscapes}, and ImageNet \citep{deng2009imagenet}, which have fueled model-centric progress in deep learning through extensive experimentation and architectural innovation \citep{whang2023data, zha2023data, sun2017revisiting}.

However, models in production demand more than benchmark-style labeling: (1) speed: Cityscapes reports 1.5 hours of expert annotation per high-resolution urban scene~\citep{cordts2016cityscapes}, and MS-COCO took over a year and a half of collaborative effort~\citep{Lin2014Microsoft}; (2) flexibility: dense datasets are hard to audit or extend, since adding a new class means revisiting every training image to insert the label; and (3) quality: dense annotation forces decisions on every pixel (including the ambiguous ones), introducing annotator biases (or errors) into the training signal. Remote sensing sharpens this problem because models trained on one sensor (multispectral, hyperspectral, SAR), platform (satellites, UAVs), or geography rarely transfer to another, so nearly every problem requires its own dataset~\citep{Rottensteiner2014, Wang2021, de2022panoptic, Rahnemoonfar2023}.

Cost-reduction strategies fall into three families, none of which removes the human from the loop. Foundation models (SAM~\citep{kirillov2023sam}, CLIPSeg~\citep{luddecke2022image}) segment from prompts but do not assign class labels in specialized taxonomies, and even when used as labeling shortcuts they produce imprecise boundaries and miss small objects in medical imaging~\citep{huang2024medical}, fine-structure scenes~\citep{ke2023hqsam}, and remote sensing~\citep{osco2023segment}, limitations that the SAM2 report~\citep{ravi2024sam2} itself acknowledges. Few-shot methods~\citep{shaban2017one, boudiaf2021fewshot} require meta-training on similar tasks, which is rare in specialized domains. The third family iterates on the target dataset by reading model predictions: weak supervision propagates sparse pixels through heuristics~\citep{Bearman2016, hua2021semantic, khoreva2017simple}, active learning selects by uncertainty~\citep{sener2018active, kellenberger2019half, mackowiak2018cereals}, pseudo-labeling reuses high-confidence predictions~\citep{lee2013pseudo, arazo2020pseudo}, and interactive frameworks combine human feedback with algorithmic label expansion~\citep{lenczner2022dial, li2023hal, yang2024easyseg}.

These output-reading approaches differ in who produces the mask (human, heuristic, model) but share the same input: the model's predictive distribution. Selection rules defined on that distribution cannot separate pixels where the model is wrong but confident from those where it is correct and confident: the two are indistinguishable at the level the rules operate on, yet they are where the most useful training signal sits. Closing this gap requires both a conceptual move and the infrastructure to test it end-to-end. \citet{yang2024easyseg} came closest on the conceptual side by observing confident errors in a domain adaptation setting, but their response layered more machinery on top (staged acquisition, pseudo-label densification, consistency regularization, domain-adaptation supervision) rather than testing whether the observation alone was enough. Concurrent work~\citep{csl2025} reports the same confident-overlap phenomenon but addresses it through output-space correction, staying within the same input regime. On the infrastructural side, the experiment requires annotation UX, prediction overlays, record format, mask generation, training pipeline, and reproducibility tooling to live in one environment. Existing annotation tools (CVAT, Labelbox, V7) stop at labeling, and existing training frameworks assume dense masks as input, so anyone running the experiment end-to-end would have to connect these stacks by hand at every iteration.

This paper proposes iSAGE (Iterative Sparse Annotation Guided by Expert), an integrated framework that fills both gaps. The annotator sees the current model's predictions overlaid on the image and clicks on pixels where the model is confidently wrong but the human can tell at a glance, providing supervision that no function over the model's predictive distribution can extract. The Error-Weighted Dice Loss (EWDL) amplifies the gradient at those clicks at every forward pass, and an integrated software platform hosts inspection, annotation, retraining, and dataset maintenance as a single continuous workflow. The contributions of this work are threefold:

\begin{enumerate}
    \item \textbf{The structural limit of output-reading supervision.} Six output-reading mechanisms used by current HITL frameworks (acquisition functions, propagation, pseudo-labels, consistency regularization, foundation-model labeling, domain adaptation) operate as functions over the model's predictive distribution, in which a confidently wrong pixel is indistinguishable from a confidently correct one by construction, yet these are precisely where annotated supervision contributes the largest gradient. The limit is informational rather than a parameter to tune, and iSAGE is the only one of 31 surveyed HITL frameworks (Table~\ref{tab:tab3}) operating without any such mechanism.

    \item \textbf{iSAGE: a framework whose sparse human clicks target this signal directly.} The annotator clicks on visible errors in the prediction overlay, and the model is retrained with those clicks as the only supervision. Under an adversarial budget of one labeled pixel per class per frame per iteration, iSAGE matches dense supervision within 0.13 mIoU on ISPRS Vaihingen (76.78 vs 76.65) and recovers 97.2\% of dense on BsB Aerial. Four output-reading baselines run under the identical pipeline plateau 7.4 to 14.5 pp below.

    \item \textbf{An open-source integrated platform that operationalizes the framework end-to-end.} The platform couples prediction overlay, click annotation, record persistence, and retraining into a single environment. As a scientific instrument, this coupling made the four-baseline falsification feasible at research scale. As a deployable tool, the platform supports new domains through preprocessing and encoder adjustments rather than methodological changes; the evidence base in this paper is aerial remote sensing. Every experiment in this paper was produced on it.
\end{enumerate}

The paper is organized as follows. Section~\ref{sec:related} reviews related work. Section~\ref{sec:methods} presents the iSAGE framework and its software platform. Section~\ref{sec:experiments} describes the experimental protocols. Section~\ref{sec:results} reports the results, including the controlled comparisons against output-reading baselines. Section~\ref{sec:discussion} discusses implications and positions iSAGE in a 31-method comparison landscape. Section~\ref{sec:conclusion} concludes.

\section{Related Work}\label{sec:related}
iSAGE is positioned against three families of prior work: sparse and weak supervision, active learning, and interactive human-in-the-loop frameworks.

\subsection{Sparse and weakly-supervised semantic segmentation}

Weak supervision strategies reduce labeling burden by replacing dense masks with less detailed annotations: image-level labels~\citep{jin2022cold}, bounding boxes~\citep{teng2022structured}, scribbles~\citep{lin2016scribblesup}, class activation maps, and selected slices in volumetric data~\citep{Cicek2016}. Sparse supervision is a stricter subset, providing only a few annotated pixels per image (typically below 1-5\%~\citep{Bearman2016, liu2021one, kellenberger2019half, hua2021semantic}), and has been applied to medical imaging~\citep{gao2022segmentation, Kervadec2019, Yang2020, Liu2023}, remote sensing~\citep{hua2021semantic, Maggiolo2022, Mazhar2022, Gbodjo2021}, and natural images~\citep{Bearman2016, lin2016scribblesup}.

How these sparse annotations reach a dense training signal divides the field into two strategies. Enhancement keeps the annotations sparse but compensates through architectural choices: Conditional Random Fields impose spatial coherence on predictions, either as post-processing or integrated modules~\citep{hua2021semantic, chen2017rethinking, can2018learning, arnab2018conditional}, and custom losses weight the gradient contribution of labeled versus unlabeled pixels~\citep{gao2022segmentation, liang2022tree, wang2022pymic, belharbi2021deep, liu2021one}. Label expansion generates pseudo-labels from sparse seeds through CRF-based refinement~\citep{ren2022framework}, GAN-based generation~\citep{desai2022active}, student-teacher consistency~\citep{wang2022pymic}, and more recently foundation models such as SAM~\citep{kirillov2023sam} and CLIPSeg~\citep{luddecke2022image} that produce class-agnostic masks for downstream category assignment. Pseudo-labeling approaches typically mitigate the confirmation-bias risk through confidence thresholds, regularization, or ensemble validation~\citep{alonso2019coralseg, Lee2020, belharbi2021deep}. Whether through enhancement or expansion, the supervision signal that ultimately trains the model is generated algorithmically from the sparse seeds rather than supplied by the annotator directly.

\subsection{Active learning for semantic segmentation}

Active learning reduces annotation effort by selecting which samples or regions a human annotates next, then refining the model after each iteration~\citep{ren2021survey}. For semantic segmentation, acquisition functions score candidate samples along two main axes: uncertainty (entropy, margin, or model disagreement) and diversity/representativeness (coverage of the input distribution), often combined with task-specific signals such as loss prediction, multi-criterion scoring, simulation-before-annotation, or Bayesian decomposition of pixel-level uncertainty~\citep{yoo2019learning, yuan2021multiple, yamani2024active, lai2021joint, rangnekar2023semantic, fan2024integrating, chen2024think, ge2024esa, balentacq2024}. Recent work evaluates whether sparse annotations achieve comparable performance to dense supervision in this setting: \citet{kellenberger2019half} showed that labeling less than 0.5\% of a dataset could outperform fully supervised models in object detection when annotations are carefully selected; \citet{li2023hal} proposed HAL for few-shot segmentation with point-based annotations and iterative correction; \citet{siddiqui2020viewal} introduced ViewAL for 3D scene annotation through viewpoint-based selection; and similar principles extend to point cloud segmentation~\citep{liu2023ococ}.

A common assumption across uncertainty-based acquisition functions is that model uncertainty serves as a reliable proxy for informativeness. This assumption breaks down in two directions: a confident incorrect prediction registers no uncertainty, and a highly uncertain boundary pixel may contribute little new information about the semantic task. \citet{mukhoti2018} found that the mapping between uncertainty and correctness in Cityscapes segmentation requires careful measurement rather than being self-evident; \citet{gustafsson2020} showed that deep ensembles improve over MC-dropout but still suffer calibration error under synthetic-to-real domain shift; and a recent benchmark of deep active learning concluded that many uncertainty-based methods barely outperform random selection~\citep{munjal2025}. These findings frame the limit of output-reading acquisition: confident errors register low uncertainty by definition, so the pixels where the training signal concentrates remain indistinguishable from confidently correct ones in any function over the model's predictive distribution.

\subsection{Interactive human-in-the-loop frameworks for semantic segmentation}

Human-in-the-loop (HITL) machine learning takes two distinct forms in computer vision~\citep{amershi2014power, mosqueira2023humans, wu2022survey}. Inference-time HITL refines individual predictions through user clicks or extreme points without altering the model~\citep{xu2016deep, maninis2018deep, sofiiuk2022reviving}. Training-time HITL contributes to training supervision via labeled queries or coupled acquisition-and-correction loops, which is the focus of this section. iSAGE belongs to the latter. Inference-time methods and acquisition-only active learning lie outside its design space.

Several training-time HITL frameworks target semantic segmentation specifically. DIAL~\citep{lenczner2022dial} combines uncertainty-driven sampling with region-based refinement and direct user feedback, using user input to compensate for cases where model uncertainty misses informative pixels. HAL~\citep{li2023hal} implements interaction loops for real-time correction of segmentation masks based on user clicks, using a dual-network mechanism that consolidates sparse clicks into dense masks for training. EasySeg~\citep{yang2024easyseg} combines a See-First-Ask-Later (SFAL) acquisition strategy, which identifies obvious model errors before querying uncertain pixels, with an ISS-Net pseudo-label generator under a domain adaptation setting, reporting performance above its fully supervised baseline at a fraction of the annotation cost. Across these frameworks, sparse human input reaches a dense training signal through an algorithmic expansion step, whether patch retraining (DIAL), dual-network consolidation (HAL), or pseudo-label generation (EasySeg).

More recent frameworks extend this pattern with foundation-model pseudo-labels. ALC~\citep{alc2024} refines SAM-generated labels through human correction queries, and A2LC~\citep{a2lc2025} automates the correction decision. Because both operate on SAM-generated regions, the training signal flows through boundaries the foundation model drew. These boundaries are imprecise on specialized domains such as medical imaging~\citep{huang2024medical}, fine-structure scenes~\citep{ke2023hqsam}, and remote sensing~\citep{osco2023segment}, and the SAM2 technical report~\citep{ravi2024sam2} acknowledges the issue itself. Making SAM usable at scale additionally requires engineering for coverage and cross-tile consistency~\citep{carvalho2026samsing}. iSAGE removes this dependency: clicks land only on pixels of visually unambiguous class, and the click itself never asks the annotator for a boundary decision.

\begin{figure}[H]
    \centering
    \includegraphics[width=\textwidth]{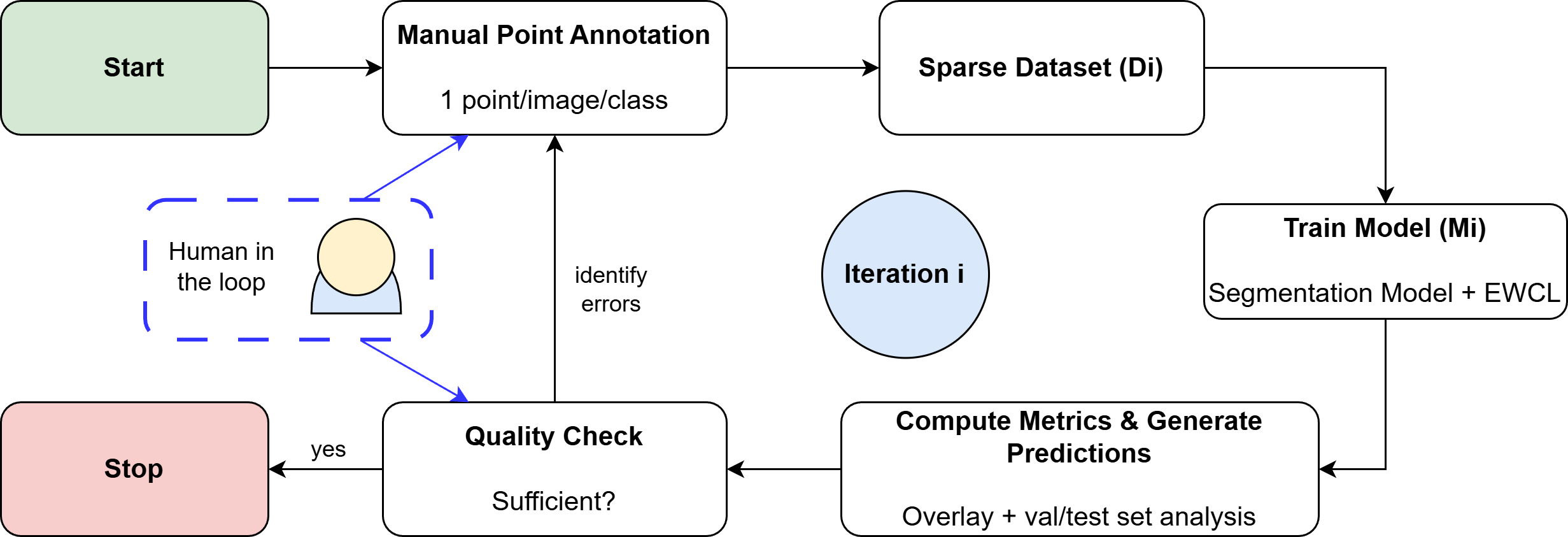}
    \caption{iSAGE workflow: iterative annotation, training with EWDL, and user-decided stopping.}
    \label{fig:fig3}
\end{figure}

\section{iSAGE Framework}\label{sec:methods}

iSAGE is a framework for semantic segmentation in which sparse human clicks drive an iterative training loop. The annotator identifies pixels where the current model is confidently wrong at each iteration. The growing record of (coordinate, class) tuples, stored as a JSON file, is converted deterministically into training masks. The Error-Weighted Dice Loss (EWDL) then trains the next iteration with those clicks amplified relative to already-correct predictions. No propagation, pseudo-labeling, or other algorithmic densification step sits between the click and the gradient. An integrated platform hosts inspection, annotation, retraining, and dataset maintenance as a single workflow. The following four subsections describe the click-based annotation, the iterative refinement loop, EWDL, and the platform. Figure~\ref{fig:fig3} overviews the iteration cycle.

\subsection{Sparse Annotations}
Sparse annotation labels only a subset of pixels in each frame. The rest are treated as ignore during training. In iSAGE, each annotation corresponds to exactly one pixel. The user clicks on a location in the image, and the software records the coordinate $(x, y)$ along with the selected class label. There is no region expansion, no bounding boxes, and no area selection. Each annotation is strictly a single pixel. In figures, annotation points may appear as enlarged markers for visibility, but the actual supervision consists only of the precise pixel coordinates. Because no propagation follows, the resulting set of $(x, y, \text{class})$ tuples constitutes the complete training supervision. The dataset produced by iSAGE is the annotation file itself, not a dense mask derived from it by an additional procedure.

The choice of a single-pixel click as the basic annotation unit is deliberate. Scribbles and sparsely-drawn polygons also reduce the total pixel count relative to dense masks, but points have a property these alternatives do not share: per-decision auditability. Each point annotation is a single (coordinate, class) tuple, individually verifiable by returning to the pixel and independently correctable at the record level (remove one JSON entry). Scribbles bundle multiple pixel-level decisions into a single primitive, so errors in part of the scribble are not easily localized or corrected. Polygons reintroduce boundary decisions, where labels are intrinsically ambiguous and annotator uncertainty becomes label noise the training procedure cannot distinguish from correct supervision. The point primitive therefore forces single-pixel granularity, which aligns with the workflow's commitment to labeling only pixels where the class is visually unambiguous. Because each click contributes exactly one pixel regardless of object size, the annotator also balances class contribution independently of spatial extent, a second advantage that follows from the same primitive.

Formally, let \( I \) be an input image and \( A: \mathbb{R}^2 \to \{0, 1, \dots, C\} \) be an annotation function that assigns class labels to pixels, where \( C \) is the number of the target class. Supervision, in sparse annotations, only considers a limited set of pixels \( S \subset \mathbb{R}^2 \) and \( A \) is undefined outside \( S \), thus:

\begin{equation}
A'(x) =
\begin{cases} 
A(x), & \text{if } x \in S, \\
-1, & \text{otherwise}.
\end{cases}
\end{equation}
\noindent Pixels with value $-1$ act as ignore mask and contribute zero gradient. This regime does not restrict what the network sees, only which pixels it is penalized on. Gradient from each labeled pixel flows through the convolutional backbone and updates all weights that influenced its prediction, and the network's spatial inductive biases (local receptive fields, translation equivariance) allow those updates to generalize to visually similar pixels elsewhere in the image. A pixel labeled as car therefore does not teach the network what that specific pixel is. It teaches the network what car pixels look like, propagating the class signal across the feature map through standard CNN machinery. Shape and boundary information emerges from this learned representation rather than from explicit boundary annotation.

For binary segmentation (\( C=1 \)), the annotation function simplifies to \( A: \mathbb{R}^2 \to \{0, 1\} \), where \( 1 \) represents the foreground and \( 0 \) represents the background. In the multiclass scenario, \( A \) assigns values from \( \{0, 1, 2, \dots, C\} \), where \( 0 \) represents the background and \( \{1, \dots, C\} \) correspond to different object classes.

\subsection{Iterative Error-Driven Refinement}
The iSAGE cycle improves the model by annotating regions where predictions are wrong. Error-focused supervision concentrates each new label on a region the current model gets wrong rather than on a randomly chosen pixel. Algorithm~\ref{alg:active_learning} describes this iterative process.

\begin{algorithm}
\caption{Iterative Error-Driven Refinement for Sparse Annotations}
\label{alg:active_learning}
\begin{algorithmic}[1]
\STATE \textbf{Input:} Initial sparse annotations \( S_0 \), number of iterations \( N \)
\STATE \textbf{Initialize:} Train segmentation model \( M_0 \) on \( S_0 \)
\FOR{\( i = 1 \) to \( N \)}
    \STATE Generate segmentation predictions \( P_i \) using model \( M_{i-1} \)
    \STATE Identify misclassified pixels (false positives and false negatives)
    \STATE Manually select new annotation points \( S'_i \) from misclassified regions
    \STATE Expand training set: \( S_{i} \gets S_{i-1} \cup S'_i \)
    \STATE Train updated model \( M_i \) on \( S_i \)
\ENDFOR
\STATE \textbf{Output:} Final trained model \( M_N \)
\end{algorithmic}
\end{algorithm}

The process starts with initial sparse annotations to train the first model. A human operator then selects new annotation points from error regions identified by the visual overlay annotation interface and tools highlighting misclassified areas. The model is retrained on the expanded training set with these new labels. The procedure is repeated for \( N \) iterations.

The experiments in this paper adopt a restrictive supervision regime as an adversarial test of the workflow: at most one pixel per class per frame per iteration. This is a worst-case setting. In practice, iSAGE places no fixed limit on annotation density, and the annotator decides how many points each frame requires. The workflow rejects any form of label propagation: no pseudo-labels, no superpixel expansion, no CRF refinement. Each annotated pixel represents exactly what the annotator selected, with no algorithmic expansion.

In iSAGE, the human's role is to supply ground truth where the model would not query for it: confident errors are indistinguishable from confident correct predictions in the model's predictive distribution, and only an external observer can identify them as wrong.

\subsection{Error-Weighted Dice Loss}

Standard loss functions treat all labeled pixels equally during training. Under sparse annotation, this can be problematic: the model receives limited supervision, and correct predictions (often the majority) dominate the gradient signal while misclassified pixels receive insufficient attention. The proposed Error-Weighted Dice Loss (EWDL) addresses this imbalance by amplifying the contribution of incorrectly predicted pixels.

\subsubsection{Formulation}

Let \( y_{gt}(x) \in \{0, 1, \dots, C-1\} \) denote the ground truth label at pixel \( x \), and let \( y_{pr}^{(c)}(x) \in [0, 1] \) be the predicted probability for class \( c \). The standard Dice loss for class \( c \) is:

\begin{equation}
\mathcal{L}_{\text{Dice}}^{(c)} = 1 - \frac{2 \sum_x y_{gt}^{(c)}(x) \cdot y_{pr}^{(c)}(x) + \epsilon}{\sum_x y_{gt}^{(c)}(x) + \sum_x y_{pr}^{(c)}(x) + \epsilon},
\end{equation}

\noindent where \( y_{gt}^{(c)}(x) \in \{0, 1\} \) is the one-hot ground truth for class \( c \), and \( \epsilon \) ensures numerical stability.

EWDL introduces a per-pixel weight based on prediction correctness. For each labeled pixel $x \in S$, a binary indicator captures whether the predicted class matches the annotator-provided ground truth:

\begin{equation}
\text{correct}(x) = \mathbb{1}\left[\arg\max_c \, y_{pr}^{(c)}(x) = y_{gt}(x)\right].
\end{equation}

\noindent The weight assigned to each pixel depends on this correctness:

\begin{equation}
w(x) =
\begin{cases}
1 & \text{if } \text{correct}(x) = 1, \\
\lambda & \text{if } \text{correct}(x) = 0,
\end{cases}
\end{equation}

\noindent where \( \lambda > 1 \) is the error penalty factor. When \( \lambda = 1 \), EWDL reduces to standard Dice loss. Higher values of \( \lambda \) increase the gradient contribution from misclassified pixels, forcing the model to focus on its mistakes.

The weighted Dice loss for class \( c \) becomes:

\begin{equation}
\mathcal{L}_{\text{EWDL}}^{(c)} = 1 - \frac{2 \sum_x w(x) \cdot y_{gt}^{(c)}(x) \cdot y_{pr}^{(c)}(x) + \epsilon}{\sum_x w(x) \cdot \left( y_{gt}^{(c)}(x) + y_{pr}^{(c)}(x) \right) + \epsilon}.
\end{equation}

\noindent For multiclass segmentation with \( C \) classes, the total loss averages across all classes:

\begin{equation}
\mathcal{L}_{\text{EWDL}} = \frac{1}{C} \sum_{c=1}^{C} \mathcal{L}_{\text{EWDL}}^{(c)}.
\end{equation}

\noindent All summations over $x$ above run over the labeled set $S$ defined above. Unlabeled pixels contribute zero weight and zero gradient, consistent with the ignore-mask behavior of standard sparse training.

\subsection{Software Platform}\label{sec:platform}

\begin{figure}[H]
    \centering
    \includegraphics[width=\textwidth]{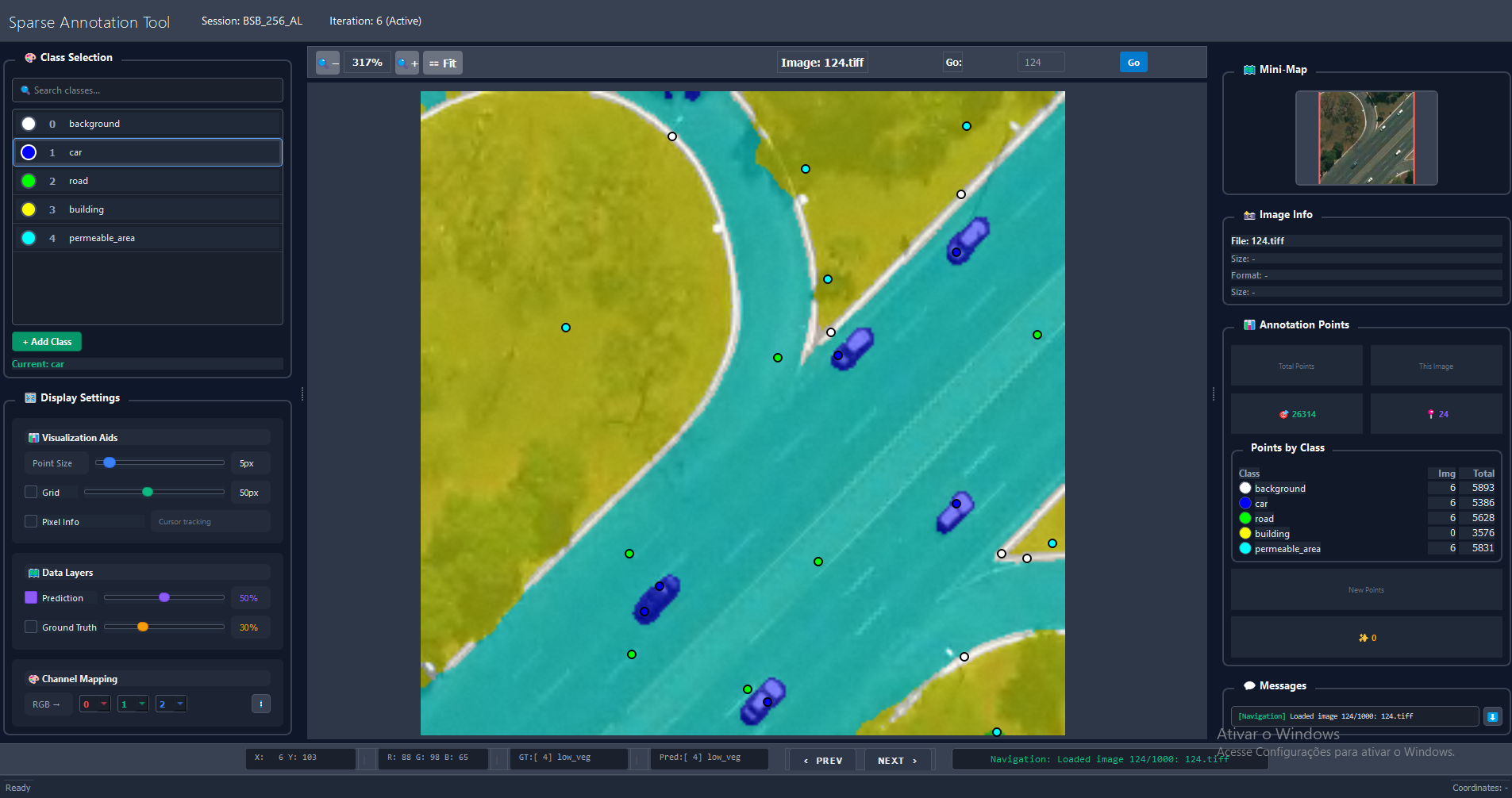}
    \caption{Annotation interface with the prediction overlay on training images.}
    \label{fig:fig4}
\end{figure}

The platform is not a convenience layer over existing tools. It is the experimental instrument that makes the alternative to output-reading supervision testable end-to-end. Without a single environment coupling prediction overlay, click annotation, record persistence, and retraining, the iterative loop fragments into ad-hoc stitching across tools, and the experiment ceases to run at research scale. Figure~\ref{fig:fig4} shows the annotation interface, with the model's prediction overlaid on the image for click-based correction.

The platform comprises four subsystems:
\begin{itemize}
    \item \textbf{Annotation interface}: hosts prediction overlay and click-driven annotation.
    \item \textbf{Record storage}: persists every decision as JSON.
    \item \textbf{Session layout}: packages each iteration as a reproducible snapshot.
    \item \textbf{Training backend}: closes the loop with EWDL-supervised retraining.
\end{itemize}
\noindent The platform is domain-agnostic by construction: any dataset with standard image formats and enumerable classes runs through the pipeline, requiring only domain-specific preprocessing and encoder choice. The training backend itself is pluggable. The default implementation uses Segmentation Models PyTorch~\citep{iakubovskii2019}, but any trainer that reads \texttt{iteration\_N/masks/} and writes the corresponding model checkpoint and predictions can replace it without touching the annotator or the record format.

The training loop does not read the JSON files directly. The annotation record is converted to dense per-image masks by a standalone mask generator, and the training dataloader consumes those masks. This separation has three consequences. First, the JSON record is a self-contained audit trail of human decisions, independent of training-specific preprocessing, and can be versioned, diffed, shared, or replayed without running the training pipeline. Second, mask generation is reproducible: given the same JSON files and the same converter, two practitioners obtain identical training masks. Third, the dataset is portable: the JSON files plus the converter are sufficient to reconstruct the supervision signal, without requiring the annotation tool or its UI state. Together these properties make the annotation record a complete, portable, auditable dataset rather than a sparse input to a densification pipeline.

\section{Experiments}\label{sec:experiments}

\subsection{Experimental setup}

All experiments use the same U-Net with EfficientNet-B7 encoder. The fixed architecture isolates annotation-strategy effects from model variability, and since iSAGE's predictions also guide annotation, architectural changes would shift point-selection trajectories across iterations. A cross-architecture validation (Section~\ref{sec:exp-bsb}) re-runs iSAGE under three additional configurations to verify the result holds across backbones. The framework is architecture-agnostic by design: any model in the Segmentation Models PyTorch library~\citep{iakubovskii2019} can replace it. Table~\ref{tab:training-config} lists the hyperparameters.

\begin{table}[!htbp]
\centering
\footnotesize
\renewcommand{\arraystretch}{1.1}
\caption{Training configuration for iSAGE experiments. Dense supervision baselines use the same architecture and optimizer settings but select the best-validation checkpoint rather than the final epoch.}
\label{tab:training-config}
\begin{tabular}{p{0.32\linewidth}p{0.55\linewidth}}
\toprule
\textbf{Setting} & \textbf{Value} \\
\midrule
Architecture & U-Net~\citep{ronneberger2015unet} \\
Encoder & EfficientNet-B7~\citep{tan2019efficientnet} \\
Optimizer & Adam ($\beta_1=0.9$, $\beta_2=0.999$) \\
Learning rate & $10^{-4}$ \\
Batch size & 10 \\
Epochs per iteration & 100 \\
Checkpoint selection & Final epoch \\
Data augmentation & Horizontal and vertical flips \\
\bottomrule
\end{tabular}
\end{table}

Performance is evaluated using IoU, F1-score, precision, and recall per class. iSAGE reports the final-epoch model, since sparse-click practitioners have no labeled validation set, so the model that ships is the one at the end of training. Dense supervision baselines use the best-validation checkpoint per standard practice, since validation labels are available in that setting.

All iSAGE experiments adopt the adversarial budget of one pixel per class per frame per iteration introduced in Section~\ref{sec:methods}, as a worst-case stress test of the workflow. \ifAnonymous The iSAGE code and configuration files are released at \url{https://github.com/anonymized/iSAGE} and archived at \url{https://doi.org/anonymized}.\else The iSAGE code and configuration files are released at \url{https://github.com/osmarluiz/iSAGE} and archived at \url{https://doi.org/10.5281/zenodo.20596185}.\fi

\subsection{Datasets}

Two complementary remote sensing datasets serve distinct roles. BsB Aerial, curated by the authors, is the controlled laboratory for comparisons against dense supervision, random selection, and alternative losses that a public benchmark cannot support. ISPRS Vaihingen is the external benchmark for comparisons against published methods and for the four output-reading falsification baselines (oracle entropy, self-training pseudo-labels, CRF-based label propagation, uniform random). The two datasets vary across geography (Brazil vs Germany), spectral composition (RGB vs IRRG), and spatial resolution (0.24m vs 0.09m).

\subsubsection{BsB Aerial}

BsB Aerial uses an aerial ortho image of Brasília, Brazil, at 0.24-meter spatial resolution and three spectral bands (Red, Green, Blue), located at approximately 15°47'32"S latitude and 47°52'10"W longitude. The image was divided into non-overlapping \mbox{$256 \times 256$} pixel frames, ensuring each annotated point appears in only one frame and preventing duplicated annotations. The study used 1{,}250 patches: 1{,}000 for training and 250 for held-out testing. The dataset was released by~\citep{de2022bounding, de2022panoptic, de2022rethinking}.

\begin{figure}[!h]
    \centering
    \includegraphics[width=\columnwidth]{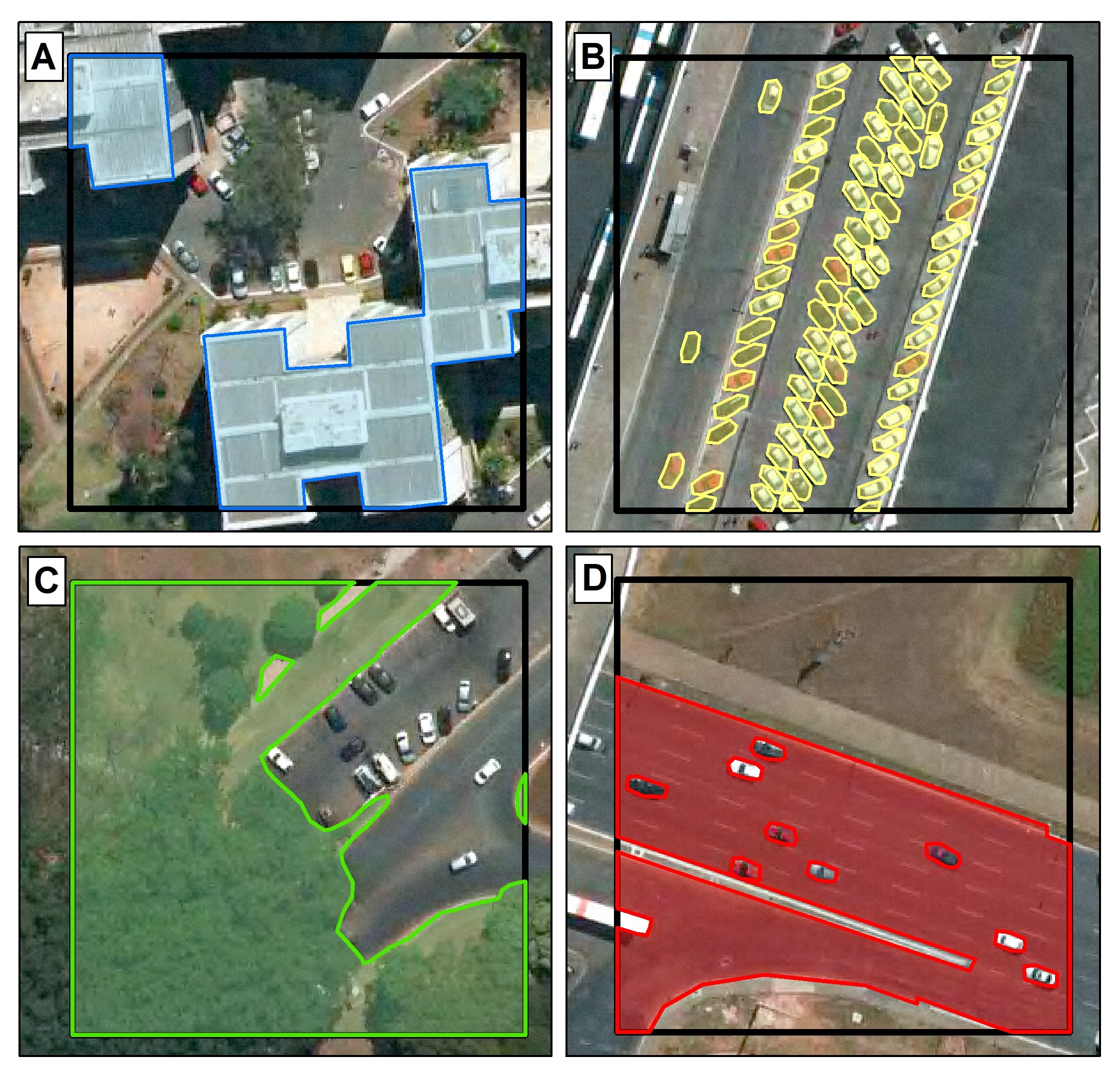}
    \caption{Examples of the four target classes: (A) buildings, (B) cars, (C) permeable areas, (D) roads.}
    \label{fig:fig2}
\end{figure}

The four BsB Aerial classes are used as an experiment in class-type isolation: each was selected to test iSAGE on a distinct combination of geometry, boundary, and texture:
\begin{itemize}
    \item \textbf{Small discrete objects} (cars): rigid geometry, sharp boundaries, low spatial frequency.
    \item \textbf{Linear connected structures} (roads): elongated, with mixed boundary regimes and frequent occlusion.
    \item \textbf{Large polygonal structures} (buildings): rectilinear, with sharp boundaries and rich interior texture.
    \item \textbf{Amorphous smooth-boundary regions} (permeable areas): irregular, with mixed textures and high intra-class variance.
\end{itemize}
Each category carries a characteristic failure mode for methods relying on model outputs or dense ground truth. Figure~\ref{fig:fig2} shows one example of each.

\subsubsection{ISPRS Vaihingen}

ISPRS Vaihingen~\citep{Rottensteiner2014} is a standard benchmark for urban semantic segmentation, with 33 aerial image tiles over Vaihingen, Germany, at 0.09m spatial resolution and three spectral bands (near-infrared, red, green). Ground truth annotations cover six semantic classes: impervious surfaces, buildings, low vegetation, trees, cars, and clutter/background.

The test split follows the same 17-tile partition used by EasySeg~\citep{yang2024easyseg}, yielding 155 non-overlapping $512 \times 512$ test patches. Training uses 1{,}000 non-overlapping $512 \times 512$ patches from the remaining 16 tiles, a 44\% reduction relative to the 1{,}784 patches used by EasySeg: this smaller training set is a deliberate constraint, not an advantage, kept consistent with iSAGE's minimum-effort regime.

Training proceeds on five classes (impervious surfaces, buildings, low vegetation, trees, cars), excluding clutter. The exclusion is not opportunistic: clutter is a catch-all category whose composition is inconsistent between training and test tiles, so single-domain learning yields zero IoU on it regardless of method. EasySeg recovers clutter only because its Potsdam domain-adaptation signal supplies external dense supervision for the class, a mechanism iSAGE deliberately avoids. The exclusion applies identically to iSAGE and to the dense baseline trained under the same protocol, so it cannot favor iSAGE: the match-with-dense result is a ratio in which both terms share the five-class setting.

\subsection{BsB Aerial experiments}\label{sec:exp-bsb}

The BsB Aerial dataset was annotated by a single annotator, holding annotator identity constant across the controlled comparisons (EWDL vs alternative losses, error-driven vs random selection, sparse vs dense supervision). External validation against an independent ground truth is provided by the Vaihingen experiments in Section~\ref{sec:exp-vaihingen}. The protocol comprises seven experiments:

\begin{itemize}
    \item \textbf{Binary experiments.} Four independent tasks, one per class, to isolate per-class convergence dynamics from class competition and observe how each visual category evolves across iterations. Each task runs for 5 iSAGE iterations after the initial seed.

    \item \textbf{Multiclass experiment.} Joint training over all four classes plus background, under the same 5-iteration protocol, to test whether multi-class supervision changes per-class behavior relative to the binary baselines.

    \item \textbf{Dense supervision upper bound.} Models trained with dense ground-truth masks under the same architecture, providing the performance ceiling iSAGE is compared against.

    \item \textbf{Random selection baseline.} The iterative protocol with randomly selected sparse annotations in place of error-driven selection, controlling for whether iSAGE's gain comes from sparsity per se or from error-targeting.

    \item \textbf{Alternative loss functions.} EWDL compared against Binary Cross-Entropy (BCE), Cross Entropy (CE), Dice, and Focal losses on the iter-5 final annotation set, testing whether EWDL's error-weighting actually contributes to iSAGE's performance or a standard loss suffices.

    \item \textbf{EWDL hyperparameter sensitivity.} The error-penalty factor $\lambda$ varied over $\{1, 2, 5, 10, 20\}$ on the iter-5 set, multi-seed, to probe the loss's stability around the chosen $\lambda=5$.

    \item \textbf{Cross-architecture validation.} Four model configurations (U-Net + EfficientNet-B7, U-Net + ResNet-101, DeepLabV3+ + ResNet-50, SegFormer + MiT-B2) paired with matched dense-supervision baselines, covering three decoder families (encoder-decoder, atrous spatial pyramid, hierarchical transformer), to confirm iSAGE's results do not depend on the specific backbone.
\end{itemize}

\subsection{ISPRS Vaihingen experiments}\label{sec:exp-vaihingen}

The Vaihingen experiments comprise two groups: an external benchmarking block where iSAGE is run on Vaihingen and compared against published methods, and four output-reading baselines that re-run the iSAGE protocol with automated acquisition mechanisms in place of the human annotator.

\subsubsection{External benchmarking}

\begin{itemize}
    \item \textbf{iSAGE on Vaihingen.} The complete iSAGE pipeline (sparse seed annotations, error-driven refinement over 5 iSAGE iterations, EWDL training) is applied to Vaihingen under the same per-iteration budget used throughout (one labeled pixel per predicted class per frame), accumulating to at most six labeled pixels per class per frame over the five iterations plus the seed. Performance is compared against published methods on EasySeg's~\citep{yang2024easyseg} 17-tile test partition.
\end{itemize}

\subsubsection{Output-reading baselines}

Four automated acquisition mechanisms are re-run under iSAGE's protocol on Vaihingen, representing the dominant automated alternatives to human-in-the-loop supervision: uncertainty-based active learning~\citep{ren2021survey, mukhoti2018}, self-training with pseudo-labels~\citep{lee2013pseudo, arazo2020pseudo}, CRF-based label propagation~\citep{krahenbuhl2011densecrf}, and uniform random sampling as a control. The goal is to test whether any of these can match iSAGE's human-targeted signal. All four share iSAGE's model, training schedule, and iter-0 seed annotations. Oracle entropy and uniform random keep the same per-iteration budget cap as iSAGE (one labeled pixel per predicted class per frame). Pseudo-labeling and CRF propagation have no per-iteration cap and expand the supervision set automatically across iterations. To rule out budget calibration and confidence-threshold calibration as the lever, the oracle is additionally evaluated at 10$\times$, 50$\times$, and 100$\times$ the per-class budget, and pseudo-labeling at confidence thresholds 0.90, 0.95, and 0.99.

\begin{itemize}
    \item \textbf{Oracle entropy~\citep{ren2021survey, mukhoti2018}}: at each iteration, the highest-entropy pixel per predicted class per frame is selected, with ground-truth labels assigned at the selected coordinates. Tests whether the strongest possible uncertainty-driven acquisition, given ground-truth answers at every query, can match iSAGE. The budget sweep at 10$\times$, 50$\times$, and 100$\times$ labels per class per frame reaches up to 0.95\% of training pixels and tests whether budget alone closes the gap to iSAGE.

    \item \textbf{Pseudo-labeling~\citep{lee2013pseudo, arazo2020pseudo}}: at each iteration, pixels whose predicted-class confidence exceeds a fixed threshold are converted to pseudo-labels and added to the supervision set. Tests whether self-confidence can expand sparse seeds into adequate supervision without human input. The threshold sweep at 0.90, 0.95, and 0.99 tests whether confidence calibration is the lever that closes the gap.

    \item \textbf{CRF-based label propagation~\citep{krahenbuhl2011densecrf}}: at each iteration, the model's softmax outputs are refined by a DenseCRF with a Gaussian spatial pairwise term and a bilateral color-position pairwise term (default hyperparameters, five mean-field iterations) before a confidence threshold matching the strongest pseudo-labeling configuration in the threshold sweep above is applied. Tests whether spatial smoothing of model outputs adds value on top of the best expansion baseline, isolating smoothing from threshold calibration.

    \item \textbf{Uniform random}: a control baseline that selects one pixel per predicted class per frame uniformly at random, with ground-truth labels assigned at the coordinates. Tests whether any acquisition structure outperforms the simplest sampling.
\end{itemize}

\section{Results}\label{sec:results}

Results follow the structure of the experimental protocol (Section~\ref{sec:experiments}). The BsB Aerial subsection reports iSAGE performance on the binary and multiclass tasks, against the random and dense reference baselines, and across loss, $\lambda$, and architecture ablations. The ISPRS Vaihingen subsection compares iSAGE against published state-of-the-art methods on the shared 17-tile test partition and reports iteration trajectories for four output-reading baselines (oracle entropy, pseudo-labeling, CRF-based label propagation, uniform random), with the structural interpretation reserved for Section~\ref{sec:disc-structural}.

\subsection{BsB Aerial}\label{sec:bsb-results}

\begin{figure}[H]
\begin{adjustwidth}{-\extralength}{0cm}
    \centering
    \includegraphics[width=\fulllength]{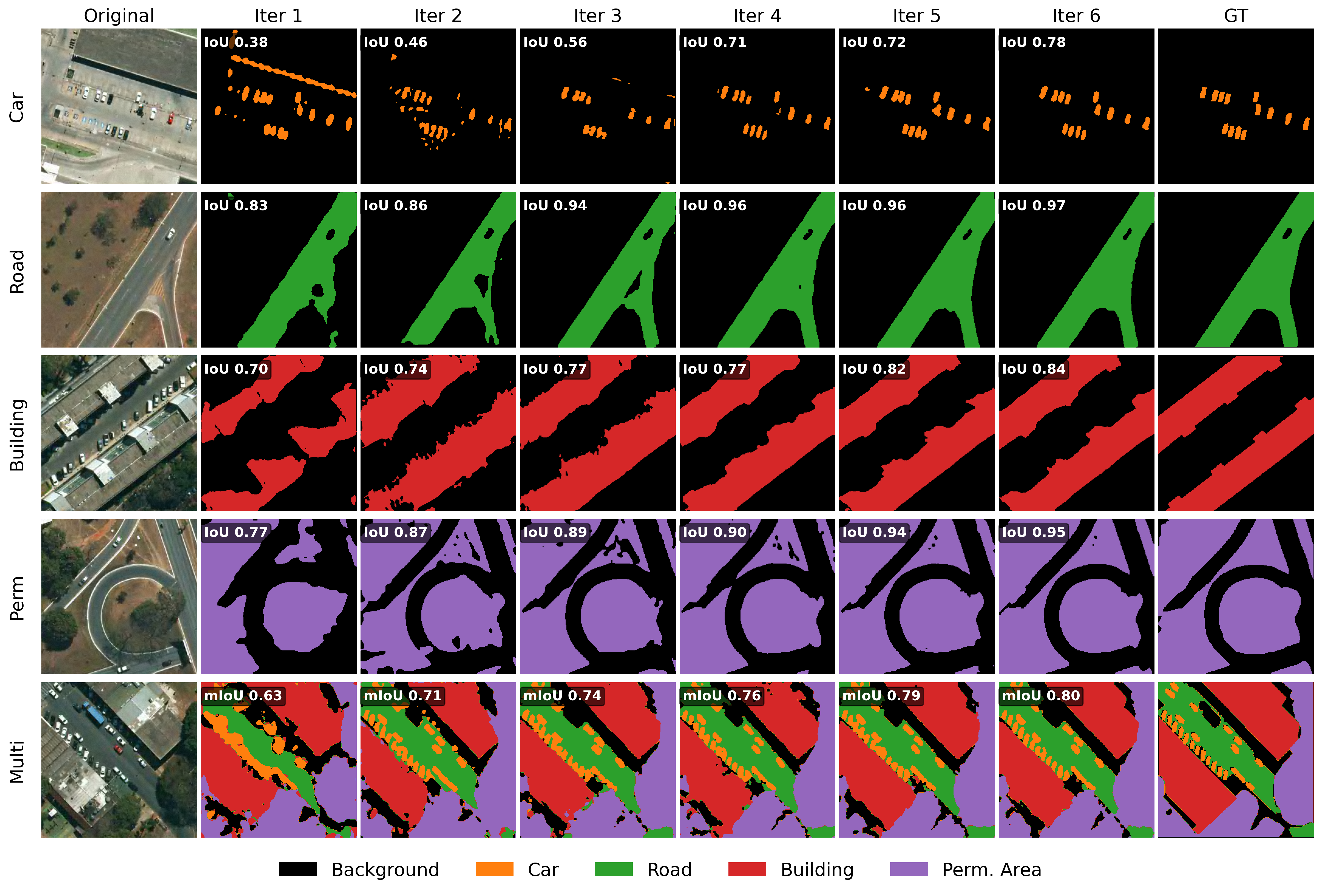}
    \caption{Visual progression of segmentation results across iSAGE iterations.}
    \label{fig:fig5}
\end{adjustwidth}
\end{figure}

On the four binary tasks, iSAGE reached final IoUs of 74.05 (car), 86.46 (road), 81.48 (building), and 88.13 (permeable area) at iteration 5 with annotation effort below 0.018\% per task, recovering 96.8\%, 96.9\%, 94.2\%, and 96.9\% of the corresponding dense IoU (Table~\ref{tab:tab1}, Figure~\ref{fig:fig5}). Cars improved by +46.6 IoU between iteration 0 and 5 and remained farthest from dense (76.52). Permeable area improved by only 9.7 IoU and started highest.

On the multiclass task, iSAGE reached 74.79\% mIoU and 84.83\% macro F1 at iteration 5 with 0.040\% cumulative annotation effort across all five classes, recovering 97.2\% of the dense ceiling (76.93\% mIoU) (Table~\ref{tab:tab2}). Per-class IoUs at iteration 5 were 88.72 (permeable area), 81.89 (building), 82.36 (road), 70.72 (car), and 50.27 (background).

\begin{figure}[H]
\begin{adjustwidth}{-\extralength}{0cm}
    \centering
    \includegraphics[width=\fulllength]{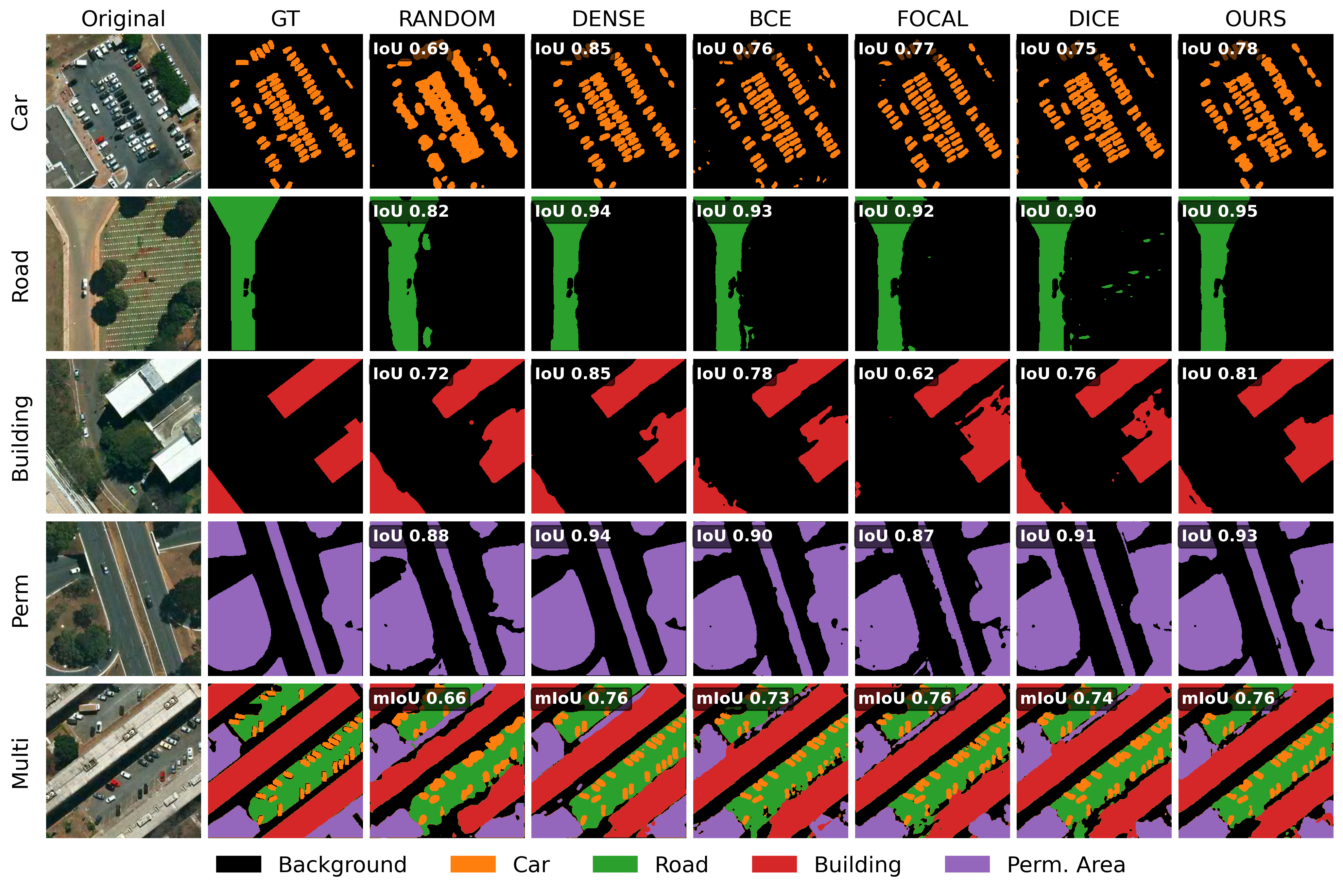}
    \caption{Qualitative comparison of final-iteration segmentation results: dense supervision, iSAGE, and random selection.}
    \label{fig:fig6}
\end{adjustwidth}
\end{figure}

On the binary tasks, random selection reached final IoUs of 39.74 (car), 77.07 (road), 73.06 (building), and 85.05 (permeable area), with gaps to iSAGE of -34.3, -9.4, -8.4, and -3.1 IoU respectively (Table~\ref{tab:tab1}, Figure~\ref{fig:fig6}). Random selection saturated the budget cap on every frame, so its cumulative effort matched or exceeded iSAGE's on three of the four classes.

On the multiclass task, random selection plateaued at 69.60\% mIoU at iteration 5, 5.19 points below iSAGE at matched annotation effort (Table~\ref{tab:tab2}).

On the four binary tasks at the iter-5 cap (Table~\ref{tab:tab1}, bottom panels), EWDL led on cars and buildings while Focal led on roads and permeable area, with all EWDL vs Focal gaps within 0.5 IoU. Dice and BCE trailed both by 1-3 IoU on cars and buildings and were within 0.5 IoU of EWDL on the other two classes.

On the multiclass task, evaluated on the iter-5 annotation set (Table~\ref{tab:ablation}), $\lambda \in \{2, 5, 10\}$ landed within 0.16 mIoU of one another (74.63, 74.79, 74.66). $\lambda = 1$ (standard Dice) reached 73.37 and $\lambda = 20$ dropped to 72.18. Against alternative loss families under the same protocol (Table~\ref{tab:tab2}, bottom panel), EWDL ($\lambda=5$) at 74.79\% led Cross-Entropy (74.11\%) by 0.68 points and matched Focal ($\gamma=2$, 74.66\%) within standard deviation.

Across four backbones trained on the same iter-5 annotation set (Table~\ref{tab:arch}), iSAGE reached 72.68 (U-Net + Eff-B7), 71.26 (U-Net + R101), 69.78 (DLV3+ + R50), and 70.34 mIoU (SegFormer + MiT-B2), recovering 94.4, 93.4, 93.2, and 93.3\% of each architecture's matched dense ceiling. Absolute mIoU spread was 2.9 points. The iSAGE/dense ratio spread was 1.2 points.

\begin{table}[!htbp]
\centering
\footnotesize
\renewcommand{\arraystretch}{1.05}
\caption{Binary segmentation results on BsB Aerial. Per-class IoU, Precision, Recall, F1-score, and annotation effort (\%).}
\label{tab:tab1}
\begin{adjustwidth}{-\extralength}{0cm}
\adjustbox{max width=\fulllength}{
\begin{tabular}{l*{4}{ccccc}}
\toprule
\textbf{Setting} & \multicolumn{5}{c}{\textbf{Car}} & \multicolumn{5}{c}{\textbf{Road}} & \multicolumn{5}{c}{\textbf{Building}} & \multicolumn{5}{c}{\textbf{Permeable Area}} \\
\cmidrule(lr){2-6} \cmidrule(lr){7-11} \cmidrule(lr){12-16} \cmidrule(lr){17-21}
 & IoU & Prec. & Recall & F1 & Effort & IoU & Prec. & Recall & F1 & Effort & IoU & Prec. & Recall & F1 & Effort & IoU & Prec. & Recall & F1 & Effort \\
\midrule

\multicolumn{21}{c}{\textbf{iSAGE (EWDL)}} \\
\midrule
Iter 0 & 27.44 & 30.69 & 72.20 & 43.07 & 0.0029 & 68.73 & 76.59 & 87.00 & 81.47 & 0.0030 & 54.98 & 60.52 & 85.71 & 70.95 & 0.0024 & 78.43 & 81.25 & 86.29 & 83.69 & 0.0030 \\
Iter 1 & 39.12 & 44.04 & 77.79 & 56.24 & 0.0058 & 73.00 & 87.21 & 81.75 & 84.39 & 0.0059 & 67.99 & 73.26 & 90.43 & 80.95 & 0.0048 & 81.34 & 82.12 & 90.57 & 86.12 & 0.0059 \\
Iter 2 & 61.09 & 73.32 & 78.55 & 75.85 & 0.0087 & 81.77 & 89.92 & 90.02 & 89.97 & 0.0086 & 76.70 & 82.40 & 91.73 & 86.81 & 0.0072 & 86.06 & 85.57 & \underline{93.20} & 89.21 & 0.0085 \\
Iter 3 & 71.38 & 84.15 & 82.46 & 83.30 & 0.0116 & 83.35 & 89.63 & 92.25 & 90.92 & 0.0110 & 80.26 & 87.50 & 90.65 & 89.05 & 0.0096 & 87.36 & 86.02 & 93.13 & 89.43 & 0.0106 \\
Iter 4 & 73.12 & \underline{86.44} & 82.59 & 84.47 & 0.0145 & 85.24 & \underline{92.68} & 91.39 & 92.03 & 0.0128 & 80.66 & 86.43 & \underline{92.36} & 89.30 & 0.0121 & 87.67 & 86.41 & 92.77 & 89.47 & 0.0122 \\
Iter 5 & \underline{74.05} & 86.02 & \underline{84.18} & \underline{85.09} & 0.0174 & \underline{86.46} & 92.36 & \underline{93.12} & \underline{92.74} & 0.0140 & \underline{81.48} & \underline{88.62} & 91.01 & \underline{89.80} & 0.0145 & \underline{88.13} & \underline{86.68} & 93.06 & \underline{89.70} & 0.0130 \\
\midrule

\multicolumn{21}{c}{\textbf{Iterative Random Selection}} \\
\midrule
Iter 0 & 31.49 & 32.73 & 89.23 & 47.89 & 0.0029 & 70.96 & 82.66 & 83.36 & 83.01 & 0.0029 & 58.00 & 69.52 & 77.78 & 73.42 & 0.0023 & 79.25 & 87.53 & 89.34 & 88.43 & 0.0030 \\
Iter 1 & 33.74 & 34.91 & 90.98 & 50.46 & 0.0058 & 74.66 & 82.71 & 88.46 & 85.49 & 0.0058 & 62.98 & 73.73 & 81.20 & 77.28 & 0.0049 & 83.49 & 87.87 & 94.37 & 91.00 & 0.0060 \\
Iter 2 & 37.17 & 38.22 & 93.13 & 54.19 & 0.0087 & 74.63 & 81.83 & 89.45 & 85.47 & 0.0089 & 66.73 & 73.46 & 87.91 & 80.04 & 0.0074 & 83.83 & 88.48 & 94.10 & 91.20 & 0.0091 \\
Iter 3 & 35.48 & 36.13 & \underline{95.21} & 52.38 & 0.0117 & 76.44 & \underline{83.09} & 90.52 & 86.65 & 0.0115 & 70.36 & 77.08 & 88.98 & 82.60 & 0.0100 & 84.62 & 88.57 & \underline{94.99} & 91.67 & 0.0121 \\
Iter 4 & 38.69 & 39.84 & 93.07 & 55.80 & 0.0145 & \underline{77.07} & 83.05 & \underline{92.60} & \underline{87.05} & 0.0146 & 70.18 & 78.14 & 87.32 & 82.48 & 0.0125 & \underline{85.13} & 89.31 & 93.17 & \underline{91.92} & 0.0151 \\
Iter 5 & \underline{39.74} & \underline{40.68} & 94.49 & \underline{56.88} & 0.0174 & 77.04 & 83.05 & 91.42 & 86.83 & 0.0171 & \underline{73.06} & \underline{78.87} & \underline{90.85} & \underline{84.44} & 0.0150 & 85.05 & \underline{90.70} & 93.17 & \underline{91.92} & 0.0181 \\
\midrule

\multicolumn{21}{c}{\textbf{Alternative Loss Functions (Final Iteration)}} \\
\midrule
BCE   & 66.96 & 79.26 & 81.18 & 80.21 & 0.0117 & \underline{84.57} & \underline{93.27} & 90.07 & \underline{91.64} & 0.0136 & \underline{77.45} & \underline{87.94} & 86.66 & \underline{87.29} & 0.0116 & \underline{86.11} & 93.33 & 91.75 & \underline{92.53} & 0.013 \\
Focal & \underline{69.13} & \underline{82.42} & 81.09 & \underline{81.75} & 0.0117 & 83.52 & 93.23 & 88.91 & 91.02 & 0.0136 & 76.05 & 85.79 & 87.02 & 86.40 & 0.0116 & 84.85 & \underline{94.68} & 89.09 & 91.80 & 0.013 \\
Dice  & 64.86 & 74.04 & \underline{83.94} & 78.68 & 0.0117 & 83.83 & 90.89 & \underline{91.52} & 91.20 & 0.0136 & 77.15 & 84.78 & \underline{89.55} & 87.10 & 0.0116 & 85.94 & 91.58 & \underline{93.31} & 92.44 & 0.013 \\
\midrule

\multicolumn{21}{c}{\textbf{Dense Supervision}} \\
\midrule
Dense & \textbf{\textit{76.52}} & \textbf{\textit{88.32}} & \textbf{\textit{85.14}} & \textbf{\textit{86.70}} & \textbf{100.0} & \textbf{\textit{89.27}} & \textbf{\textit{95.89}} & \textbf{\textit{92.83}} & \textbf{\textit{94.33}} & \textbf{100.0} & \textbf{\textit{86.46}} & \textbf{\textit{93.91}} & \textbf{\textit{91.59}} & \textbf{\textit{92.74}} & \textbf{100.0} & \textbf{\textit{90.96}} & \textbf{\textit{94.08}} & \textbf{\textit{96.48}} & \textbf{\textit{95.26}} & \textbf{100.0} \\
\bottomrule
\end{tabular}
}
\end{adjustwidth}
\end{table}

\begin{table}[!htbp]
\centering
\footnotesize
\renewcommand{\arraystretch}{1.05}
\caption{Multiclass segmentation results on BsB Aerial. Macro metrics and per-class IoU (\%).}
\label{tab:tab2}
\begin{adjustwidth}{-\extralength}{0cm}
\adjustbox{max width=\fulllength}{
\begin{tabular}{lccccccccccc}
\toprule
& \multicolumn{5}{c}{\textbf{Macro Metrics (\%)}} & \multicolumn{5}{c}{\textbf{IoU per Class (\%)}} \\
\cmidrule(lr){2-6} \cmidrule(lr){7-11}
\textbf{Setting} & \textbf{mIoU} & \textbf{mPrecision} & \textbf{mRecall} & \textbf{mF1} & \textbf{Effort} & \textbf{Background} & \textbf{Car} & \textbf{Road} & \textbf{Building} & \textbf{Perm. Area} \\
\midrule
\multicolumn{11}{c}{\textbf{iSAGE (EWDL)}} \\
\midrule
Iter 0 & 58.38 & 71.63 & 77.08 & 72.97 & 0.0068 & 38.46 & 47.08 & 68.90 & 59.71 & 77.74 \\
Iter 1 & 67.89 & 77.91 & 83.13 & 79.81 & 0.0135 & 45.64 & 56.30 & 78.18 & 73.08 & 86.23 \\
Iter 2 & 72.05 & 82.34 & 83.41 & 82.86 & 0.0202 & 47.45 & 67.36 & 79.72 & 78.05 & 87.69 \\
Iter 3 & 73.19 & 83.04 & 84.79 & 83.80 & 0.0270 & 47.89 & 68.03 & 81.37 & 80.52 & 88.13 \\
Iter 4 & 74.14 & 83.41 & 85.64 & 84.36 & 0.0337 & 49.17 & 69.86 & 81.81 & 81.48 & 88.38 \\
Iter 5 & \underline{74.79} & \underline{84.07} & \underline{85.72} & \underline{84.83} & 0.0404 & \underline{50.27} & \underline{70.72} & \underline{82.36} & \underline{81.89} & \underline{88.72} \\
\midrule
\multicolumn{11}{c}{\textbf{Iterative Random Selection}} \\
\midrule
Iter 0 & 56.24 & 69.63 & 75.26 & 70.99 & 0.0015 & 37.73 & 46.06 & 66.19 & 54.88 & 76.34 \\
Iter 1 & 64.73 & 75.99 & 81.27 & 77.56 & 0.0136 & 41.65 & 53.89 & 74.51 & 71.68 & 81.95 \\
Iter 2 & 66.67 & 77.49 & 82.35 & 79.04 & 0.0205 & 43.23 & 57.50 & 75.31 & 73.76 & 83.55 \\
Iter 3 & 67.63 & 78.16 & 82.69 & 79.75 & 0.0273 & 43.74 & 59.21 & 76.06 & 75.07 & 84.08 \\
Iter 4 & 68.41 & 78.77 & \underline{83.25} & 80.34 & 0.0342 & 44.81 & 60.15 & 76.72 & 75.55 & 84.84 \\
Iter 5 & \underline{69.60} & \underline{79.39} & \underline{83.25} & \underline{80.83} & 0.0410 & \underline{45.02} & \underline{61.73} & \underline{77.08} & \underline{76.33} & \underline{85.36} \\
\midrule
\multicolumn{11}{c}{\textbf{Dense Supervision}} \\
\midrule
Dense & 76.93 & 87.04 & 86.23 & 86.47 & 100.0 & 53.72 & 73.14 & 83.70 & 84.65 & 90.73 \\
\midrule
\multicolumn{11}{c}{\textbf{Alternative Loss Functions (Final Iteration)}} \\
\midrule
Dice ($\lambda=1$) & 73.37 & 82.78 & 85.12 & 83.81 & 0.0404 & 48.15 & 68.66 & 81.63 & 80.28 & 88.15 \\
Cross-Entropy & 74.11 & 83.35 & 85.53 & 84.32 & 0.0404 & 48.94 & 69.52 & \underline{82.29} & 81.38 & 88.43 \\
Focal ($\gamma=2$) & \underline{74.66} & \underline{83.85} & \underline{85.82} & \underline{84.74} & 0.0404 & \underline{50.26} & \underline{70.24} & 82.07 & \underline{82.04} & \underline{88.69} \\
\bottomrule
\end{tabular}
}
\end{adjustwidth}
\end{table}

\begin{table}[!htbp]
\centering
\caption{EWDL hyperparameter sensitivity on BsB Aerial multiclass. Mean mIoU and per-class IoU (\%) across 5 seeds under varying error-penalty $\lambda$ ($\lambda=1$ recovers standard Dice).}
\label{tab:ablation}
\begin{tabular}{lcccccc}
\toprule
$\lambda$ & \textbf{mIoU} & \textbf{Car} & \textbf{Road} & \textbf{Building} & \textbf{Perm.} & \textbf{BG} \\
\midrule
1 (= standard Dice) & 73.37 & 68.66 & 81.63 & 80.28 & 88.15 & 48.15 \\
2 & 74.63 & 70.38 & 82.23 & \underline{82.03} & 88.53 & 49.97 \\
\textbf{5 (default)} & \underline{74.79} & 70.72 & 82.36 & 81.89 & \underline{88.72} & \underline{50.27} \\
10 & 74.66 & 71.29 & \underline{82.51} & 81.40 & 88.60 & 49.50 \\
20 & 72.18 & \underline{71.68} & \underline{82.51} & 73.84 & 88.55 & 44.35 \\
\bottomrule
\end{tabular}
\end{table}

\begin{table}[!htbp]
\centering
\small
\caption{Cross-architecture validation on BsB Aerial multiclass. mIoU and per-class IoU (\%) for iSAGE and matched-architecture dense baselines.}
\label{tab:arch}
\begin{adjustwidth}{-\extralength}{0cm}
\adjustbox{max width=\fulllength}{
\begin{tabular}{llcccccccc}
\toprule
 & & \multicolumn{6}{c}{\textbf{iSAGE (\%)}} & \textbf{Dense} & \textbf{iSAGE/} \\
\cmidrule(lr){3-8}
\textbf{Architecture} & \textbf{Encoder} & \textbf{mIoU} & \textbf{Car} & \textbf{Road} & \textbf{Build.} & \textbf{Perm.} & \textbf{BG} & \textbf{mIoU} & \textbf{Dense} \\
\midrule
U-Net (baseline) & Eff-B7 & \underline{72.68} & \underline{69.20} & 79.80 & 79.91 & \underline{89.26} & \underline{45.21} & \underline{77.03} & \underline{94.4\%} \\
U-Net      & R101   & 71.26 & 66.38 & \underline{79.84} & 78.64 & 88.20 & 43.23 & 76.30 & 93.4\% \\
DLV3+      & R50    & 69.78 & 58.95 & 78.69 & 79.92 & 87.34 & 44.02 & 74.85 & 93.2\% \\
SegFormer  & MiT-B2 & 70.34 & 59.40 & 78.17 & \underline{80.88} & 88.27 & 44.98 & 75.41 & 93.3\% \\
\bottomrule
\end{tabular}
}
\end{adjustwidth}
\end{table}

\FloatBarrier

\subsection{ISPRS Vaihingen}\label{sec:vaihingen-results}

\subsubsection{External benchmarking}

iSAGE reached 76.78\% mIoU on the ISPRS Vaihingen five-class benchmark at iteration 5 with 29{,}052 labeled pixels (0.011\% of training pixels), matching the dense baseline trained under iSAGE's protocol (76.65\%) within 0.13 points (Table~\ref{tab:vaihingen}). Against published weakly-supervised methods, iSAGE led EasySeg by 3.95 points (76.78 vs 72.83), D2ADA by 5.03 (vs 71.75), ILM-ASSL by 6.36 (vs 70.42), and RIPU by 6.87 (vs 69.91), while consuming about one-third of EasySeg's annotation budget and training on 44\% fewer patches. Competitor numbers are as reported in EasySeg's evaluation~\citep{yang2024easyseg} rather than re-run under iSAGE's protocol; the load-bearing internal control is the dense baseline trained under iSAGE's protocol (76.65\%), reported in the same table. The protocol asymmetry described in Section~\ref{sec:experiments} (iSAGE final-epoch vs dense best-validation) is verified to have negligible impact: the gap between iSAGE's final-epoch and best-validation checkpoints on Vaihingen is within 0.03 mIoU under the learning-rate decay schedule used here.

\begin{table}[!htbp]
\centering
\small
\caption{Comparison on ISPRS Vaihingen considering per-class IoU (\%) and labeling cost. Weakly-supervised competitor numbers from~\citet{yang2024easyseg}.}
\label{tab:vaihingen}
\begin{tabular}{llcccccc}
\toprule
\textbf{Method} & \textbf{Cost} & \textbf{mIoU} & \textbf{Imperv.} & \textbf{Build.} & \textbf{Tree} & \textbf{Car} & \textbf{Low Veg.} \\
\midrule
\multicolumn{8}{c}{\textit{Unsupervised Domain Adaptation (no target labels)}} \\
\midrule
No Adaptation & 0\% & 30.19 & 35.24 & 42.23 & 43.57 & 1.39 & 28.50 \\
ADVENT~\citeyearpar{vu2019advent} & 0\% & 40.84 & 53.82 & 56.59 & 49.21 & 24.46 & 20.12 \\
CLAN~\citeyearpar{luo2019clan} & 0\% & 46.40 & 63.38 & 61.78 & 55.03 & 28.98 & 22.83 \\
\midrule
\multicolumn{8}{c}{\textit{Active Domain Adaptation}} \\
\midrule
MADAv2~\citeyearpar{ning2023madav2} & 1\% & 50.62 & 65.73 & 59.94 & 54.26 & 21.57 & 51.58 \\
RIPU~\citeyearpar{xie2022ripu} & 0.015\% & 69.91 & 80.09 & 86.17 & 65.02 & 53.02 & 65.26 \\
ILM-ASSL~\citeyearpar{guan2023ilmassl} & 1\% & 70.42 & 81.03 & 87.28 & 64.36 & 57.13 & 62.32 \\
D2ADA~\citeyearpar{wu2022d2ada} & 1\% & 71.75 & 80.71 & 87.62 & 66.69 & 57.58 & 66.17 \\
EasySeg~\citeyearpar{yang2024easyseg} & 0.015\% & 72.83 & 81.62 & 88.40 & 69.16 & 57.90 & 67.25 \\
\midrule
\multicolumn{8}{c}{\textit{Supervised Learning}} \\
\midrule
Fully Supervised & 100\% & 76.65 & \textbf{85.25} & \textbf{88.75} & 70.80 & 69.20 & 69.28 \\
\midrule
\textbf{iSAGE} & 0.011\% & \textbf{76.78} & 84.02 & 88.40 & \textbf{71.63} & \textbf{70.10} & \textbf{69.75} \\
\bottomrule
\end{tabular}
\end{table}

The per-class breakdown at iteration 5 (Table~\ref{tab:vaihingen}) concentrates the iSAGE-vs-published gap on cars: iSAGE reached 70.10\% car IoU against 57.90 for EasySeg (the best published competitor on this class). Across the other four classes, iSAGE led EasySeg by 2.40 on impervious (84.02 vs 81.62), 2.47 on trees (71.63 vs 69.16), and 2.50 on low vegetation (69.75 vs 67.25), and tied on buildings (88.40).

\subsubsection{Output-reading baselines}

All four baselines share iSAGE's iter-0 seed annotations (0.011\% of pixels, the same model bootstrap used by iSAGE) and the same training schedule and architecture. Oracle entropy and uniform random keep the same per-iteration budget cap as iSAGE (one labeled pixel per predicted class per frame, ground-truth labels assigned at the queried coordinates), with the oracle additionally evaluated at 10$\times$, 50$\times$, and 100$\times$ that budget. Pseudo-labeling and CRF propagation expand the supervision set across iterations by adding every pixel whose predicted-class confidence exceeds the threshold, so by iteration 5 the training mask covers approximately 90\% of the training pixels rather than 0.011\%. Pseudo-labeling is evaluated at confidence thresholds 0.90, 0.95, and 0.99. The comparison is therefore between iSAGE with 29{,}052 human-targeted pixels and the expansion baselines with roughly 230 million auto-generated pseudo-labels.

\begin{figure}[!htbp]
    \centering
    \includegraphics[width=0.85\columnwidth]{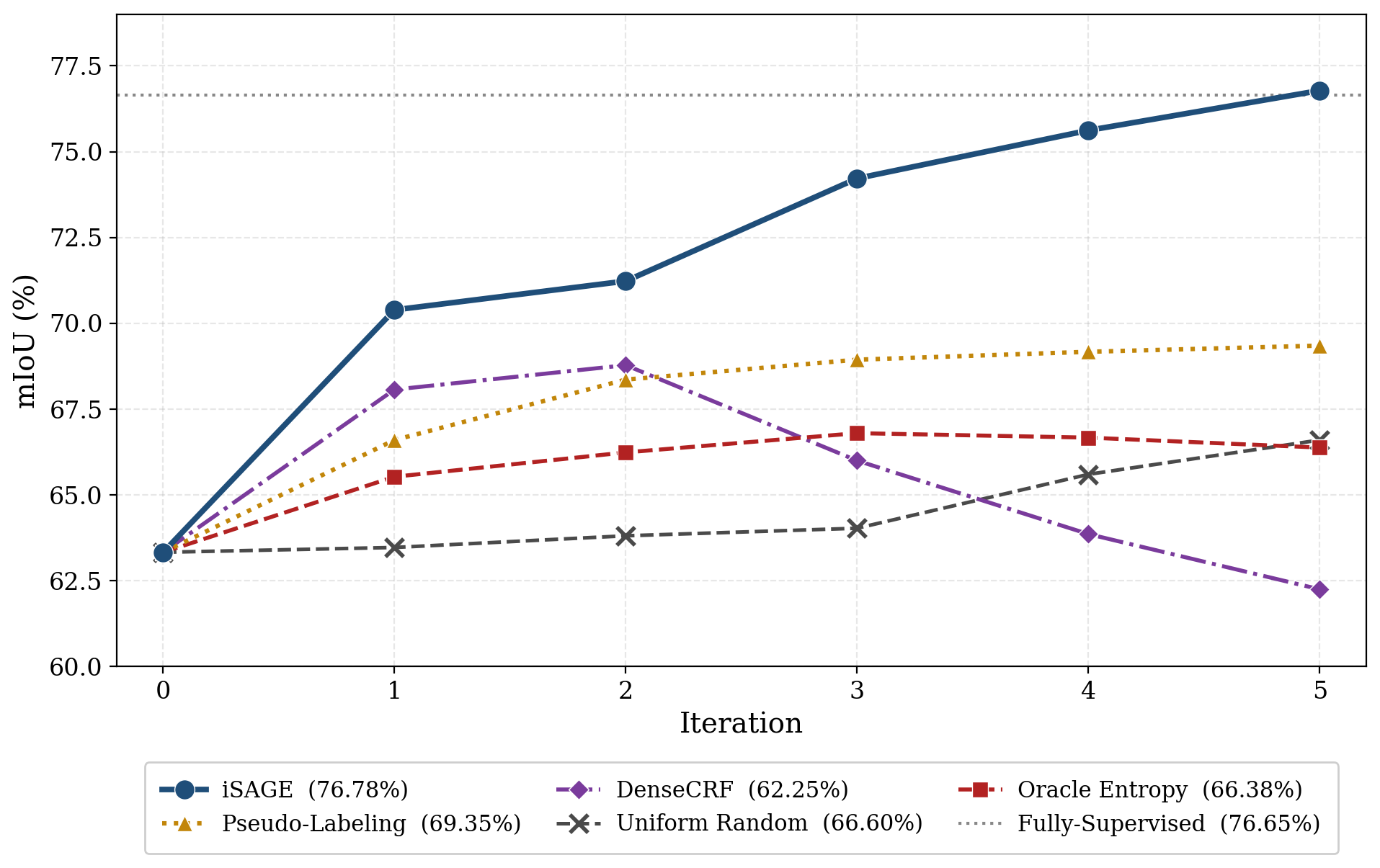}
    \caption{Per-iteration mIoU on ISPRS Vaihingen for iSAGE and the four output-reading baselines (oracle entropy, pseudo-labeling, DenseCRF label propagation, uniform random) under identical protocol conditions. The fully-supervised dense baseline trained under iSAGE's protocol is shown for reference.}
    \label{fig:miou-progression}
\end{figure}

Oracle entropy (Figure~\ref{fig:miou-progression}) plateaus at 66.38\% mIoU by iteration 3 and stays there through iteration 5, 10.40 points below iSAGE. Raising the oracle budget to 10$\times$, 50$\times$, and 100$\times$ moves the iteration-5 plateau to 66.26\%, 67.01\%, and 67.85\%, the last reaching 0.95\% of training pixels and still 8.93 points below iSAGE. Uniform random tracks the 1$\times$ oracle closely at 66.60\% at iteration 5 (gap 0.22 points to the oracle), confirming that the model's output distribution carries no acquisition signal beyond what uniform sampling extracts. Pseudo-labeling at confidence 0.95 rises monotonically to 69.35\% at iteration 5 with per-iteration gains shrinking from +1.76 to +0.18 between consecutive rounds, plateauing 7.43 points below iSAGE even at the 230-million-pixel supervision scale. At thresholds 0.90 and 0.99 the iteration-5 plateau is 69.00\% and 69.34\%, a 0.35 pp spread across the three thresholds. CRF-based label propagation peaks at 68.78\% at iteration 2 and then descends to 66.00\%, 63.86\%, and 62.25\% at iterations 3, 4, and 5, ending 14.53 points below iSAGE and 7.10 points below its own peak (Table~\ref{tab:output-reading}).

\begin{table}[!htbp]
\centering
\small
\caption{Iteration-5 mIoU on ISPRS Vaihingen for iSAGE and the output-reading baselines under identical protocol. All baselines start from the same iter-0 seed model. ``Pixels labeled'' reports the cumulative percentage of training pixels carrying a supervision signal at iteration 5.}
\label{tab:output-reading}
\begin{tabular}{lccc}
\toprule
\textbf{Method} & \textbf{Pixels labeled (\%)} & \textbf{Iter-5 mIoU (\%)} & \textbf{Gap to iSAGE (pp)} \\
\midrule
\textbf{iSAGE}                  & 0.011               & \textbf{76.78}  & 0.00      \\
\midrule
\multicolumn{4}{c}{\textit{Uncertainty acquisition (oracle entropy budget sweep)~\citep{ren2021survey, mukhoti2018}}} \\
\midrule
Oracle entropy 1$\times$        & 0.011               & 66.38           & $-10.40$  \\
Oracle entropy 10$\times$       & 0.10                & 66.26           & $-10.52$  \\
Oracle entropy 50$\times$       & 0.48                & 67.01           & $-9.77$   \\
Oracle entropy 100$\times$      & 0.95                & 67.85           & $-8.93$   \\
\midrule
\multicolumn{4}{c}{\textit{Self-training pseudo-labels (confidence threshold sweep)~\citep{lee2013pseudo, arazo2020pseudo}}} \\
\midrule
Pseudo-labeling (0.90)          & $\sim$95            & 69.00           & $-7.78$   \\
Pseudo-labeling (0.95)          & 93.0                & 69.35           & $-7.43$   \\
Pseudo-labeling (0.99)          & 84.8                & 69.34           & $-7.44$   \\
\midrule
\multicolumn{4}{c}{\textit{CRF-based label propagation~\citep{krahenbuhl2011densecrf}}} \\
\midrule
DenseCRF (0.95)                 & 96.7                & 62.25           & $-14.53$  \\
\midrule
\multicolumn{4}{c}{\textit{Control}} \\
\midrule
Uniform random                  & 0.011               & 66.60           & $-10.18$  \\
\bottomrule
\end{tabular}
\end{table}

\FloatBarrier

\section{Discussion}\label{sec:discussion}

This section interprets the results in four parts. Section~\ref{sec:disc-structural} places iSAGE among existing frameworks and argues that its machinery-free design is principled rather than incidental, resting on a structural limit of what a model's own outputs can reveal. Section~\ref{sec:disc-findings} reports what the experiments reveal about how iSAGE behaves. Section~\ref{sec:disc-beyond-benchmark} draws out the operational and cognitive properties that follow from its minimal design. Section~\ref{sec:limitations} sets out the limitations and the directions they open.

\subsection{iSAGE in the landscape and why this position is justified}\label{sec:disc-structural}

\subsubsection{The output-reading limit}

This paper argues that the six output-reading mechanisms current frameworks layer on top of human supervision do not address what they were assumed to address: they cannot separate the pixels where the model is confidently wrong from those where it is confidently correct, because the two are indistinguishable in the model's predictive distribution by construction. iSAGE is the methodological consequence of taking that limit seriously, a sparse human click delivered directly on the image with no machinery in between.

Three theoretical regimes lie outside this argument. Ensemble disagreement across independently trained models can in principle surface a subset of confident errors when ensemble members fail in uncorrelated ways. Out-of-distribution detection scores can flag confident errors that coincide with covariate shift. A supervised error-detector trained on image features with labeled examples of model errors is a function over outputs augmented with external error supervision. All three require resources the present setting does not assume (multiple parallel training budgets, an in-distribution prior, or labeled error examples respectively), and the empirical reach of each is limited: correlated ensemble failures persist under the calibration regimes studied by \citet{gustafsson2020}, OOD scores degrade precisely when the wrong-class confidence is shape- or texture-driven rather than distribution-driven, and supervised error-detectors require the very signal whose acquisition iSAGE addresses. The structural limit therefore holds within the single-model HITL regime that the 31-method survey of Table~\ref{tab:tab3} occupies and that this paper investigates, and the rest of the argument is framed accordingly.

\subsubsection{The only framework without auxiliary machinery}

Of the 31 frameworks surveyed, iSAGE is the only one that adds no auxiliary machinery at all, and the experiments below show that this subtraction costs nothing.

Table~\ref{tab:tab3} positions iSAGE alongside 30 other methods across eight supervision mechanisms. The categorization is descriptive of mechanism types rather than a measurement of relative effectiveness. Within this survey, every iterative human-in-the-loop method incorporates at least one of the six auxiliary mechanisms (acquisition, propagation, pseudo-labels, consistency regularization, domain adaptation, or foundation-model labeling). iSAGE is the single exception.

\begin{table}[!htbp]
\centering
\scriptsize
\renewcommand{\arraystretch}{0.9}
\caption{Conceptual comparison of segmentation frameworks across supervision mechanisms. Column abbreviations: Iter., iterative execution; HIL, human-in-the-loop feedback; Acq., automated acquisition function over the model's predictive distribution; Prop., propagation heuristics; Pseudo, pseudo-labels; Consist., consistency regularization; DA, domain-adaptation source; FM, foundation-model labeling. A check mark indicates the mechanism is present.}
\label{tab:tab3}
\begin{adjustwidth}{-\extralength}{0cm}
\adjustbox{max width=\fulllength, max totalheight=0.85\textheight}{
\begin{tabular}{lccccccccc}
\toprule
\textbf{Method} & \textbf{Annotation} & \textbf{Iter.} & \textbf{HIL} & \textbf{Acq.} & \textbf{Prop.} & \textbf{Pseudo} & \textbf{Consist.} & \textbf{DA} & \textbf{FM} \\
\midrule
\textbf{iSAGE} & Sparse clicks & \checkmark & \checkmark &  &  &  &  &  &  \\
\midrule
\multicolumn{10}{c}{\textit{Weak / sparse supervision (non-iterative)}} \\
\midrule
\citet{Bearman2016} & Sparse points &  &  &  &  &  &  &  &  \\
ScribbleSup~\citeyearpar{lin2016scribblesup} & Scribbles &  &  &  & \checkmark &  &  &  &  \\
FESTA~\citeyearpar{hua2021semantic} & Scribbles &  &  &  & \checkmark &  &  &  &  \\
\citet{liu2021one} & One point &  &  &  & \checkmark &  &  &  &  \\
\midrule
\multicolumn{10}{c}{\textit{Iterative weak / self-training / semi-supervised}} \\
\midrule
Tree Energy~\citeyearpar{liang2022tree} & Scribbles & \checkmark &  &  & \checkmark & \checkmark &  &  &  \\
ScribFormer~\citeyearpar{li2024scribformer} & Scribbles & \checkmark &  &  &  & \checkmark & \checkmark &  &  \\
PyMIC~\citeyearpar{wang2022pymic} & Points/Scribbles & \checkmark &  &  &  & \checkmark & \checkmark &  &  \\
UniMatch V2~\citeyearpar{yang2025unimatch} & Unlabeled + few & \checkmark &  &  &  & \checkmark & \checkmark &  &  \\
\midrule
\multicolumn{10}{c}{\textit{Active learning / interactive (single-domain)}} \\
\midrule
\citet{desai2022active} & Points &  &  & \checkmark &  & \checkmark &  &  &  \\
DIAL~\citeyearpar{lenczner2022dial} & Clicks & \checkmark & \checkmark & \checkmark & \checkmark &  &  &  &  \\
HAL-IA~\citeyearpar{li2023hal} & Sparse points & \checkmark & \checkmark & \checkmark & \checkmark &  &  &  &  \\
ViewAL~\citeyearpar{siddiqui2020viewal} & 3D points & \checkmark &  & \checkmark &  &  &  &  &  \\
S4AL~\citeyearpar{rangnekar2023semantic} & Points & \checkmark &  & \checkmark &  & \checkmark &  &  &  \\
ESA~\citeyearpar{ge2024esa} & Clicks & \checkmark & \checkmark & \checkmark & \checkmark &  &  &  &  \\
Think Twice~\citeyearpar{chen2024think} & Points & \checkmark &  & \checkmark &  &  &  &  &  \\
Cold AL~\citeyearpar{jin2022cold} & Points & \checkmark &  & \checkmark &  &  &  &  &  \\
\citet{teng2022structured} & Points & \checkmark &  & \checkmark &  &  &  &  &  \\
Adapt. Superpixel AL~\citeyearpar{kim2023adaptive} & Superpixels & \checkmark &  & \checkmark & \checkmark &  &  &  &  \\
BalEntAcq~\citeyearpar{balentacq2024} & Sparse pixels & \checkmark &  & \checkmark &  &  &  &  &  \\
ALC~\citeyearpar{alc2024} & Clicks & \checkmark & \checkmark & \checkmark &  & \checkmark &  &  & self \\
A2LC~\citeyearpar{a2lc2025} & Clicks & \checkmark & \checkmark & \checkmark &  & \checkmark &  &  & self \\
\midrule
\multicolumn{10}{c}{\textit{Active domain adaptation}} \\
\midrule
RIPU~\citeyearpar{xie2022ripu} & Regions & \checkmark &  & \checkmark &  &  & \checkmark & \checkmark &  \\
D2ADA~\citeyearpar{wu2022d2ada} & Regions & \checkmark &  & \checkmark &  &  &  & \checkmark &  \\
ILM-ASSL~\citeyearpar{guan2023ilmassl} & Sparse points & \checkmark &  & \checkmark &  & \checkmark &  & \checkmark &  \\
EasySeg~\citeyearpar{yang2024easyseg} & Sparse points & \checkmark & \checkmark & \checkmark & \checkmark & \checkmark & \checkmark & \checkmark &  \\
\midrule
\multicolumn{10}{c}{\textit{Few-shot / zero-shot / foundation-model}} \\
\midrule
PANet~\citeyearpar{wang2019panet} & Support masks &  &  &  &  &  &  &  &  \\
RePRI~\citeyearpar{boudiaf2021fewshot} & Support masks &  &  &  &  &  &  &  &  \\
SAM~\citeyearpar{kirillov2023sam} & Prompts/clicks &  &  &  &  &  &  &  & self \\
CLIPSeg~\citeyearpar{luddecke2022image} & Text/image &  &  &  &  &  &  &  & self \\
SEEM~\citeyearpar{zou2023segment} & Prompts/text &  & \checkmark &  &  &  &  &  & self \\
\bottomrule
\end{tabular}
}
\end{adjustwidth}
\end{table}

The minimalism is informational rather than stylistic. Any selection rule defined on the model's own predictive distribution cannot separate a confidently wrong pixel from a confidently correct one: at the level of that distribution the two are identical by construction. The distinguishing information lies outside the model's outputs, and supplying it is exactly the role of the human click. Each machinery column except domain adaptation is a rule operating on that distribution in some form. Acquisition functions surface uncertainty rather than error, propagation heuristics extrapolate from the same predictions, pseudo-labeling and consistency regularization train on the model's own beliefs, and foundation-model labeling swaps the target model for a different model with the same blind spot, also inheriting that model's boundary imprecision as a second source of noise. Domain adaptation is the one column outside this argument because it addresses cross-domain transfer rather than intra-domain confident errors.

iSAGE keeps only three components, the human click, the loss, and the platform, and the annotation record itself becomes the supervision artifact. A given record uniquely determines the training signal with no propagation hyperparameters, so reported performance reflects exactly what the annotator provided.

\subsubsection{Theoretical and empirical support}

Recent theoretical work argues that the aleatoric-epistemic dichotomy is insufficiently expressive and that popular information-theoretic measures over outputs are poor estimators of what they purport to quantify~\citep{aleatoric-epistemic2024}, which provides formal backing for this structural limit beyond the specific acquisition and propagation mechanisms tested here. Bayesian acquisition signals such as MC-dropout disagreement~\citep{mukhoti2018}, deep ensembles~\citep{gustafsson2020}, and posterior-epistemic decompositions~\citep{balentacq2024} remain within the output-reading regime the structural argument bounds.

Four output-reading mechanisms confirm the prediction. Oracle-entropy acquisition, given ground-truth answers on every query at iSAGE's 0.011\% per-iteration budget, plateaus at 66.38\% mIoU and remains 10.40 points below iSAGE. Raising the oracle budget to 10$\times$, 50$\times$, and 100$\times$ labels per class per frame holds the plateau between 66.26\% and 67.85\%, the last reaching 0.95\% of training pixels and still 8.93 points below iSAGE. Uniform random selection at the 1$\times$ budget tracks the oracle within seed variance, confirming that no acquisition reading the model's predictive distribution outperforms a baseline that does not. The reason is geometric: confident errors lie in the low-entropy region of the predictive distribution, precisely where uncertainty-based acquisition does not look. An entropy oracle, even handed the correct label at every query, never selects them, which is why it performs no better than uniform random. Self-training pseudo-labels rise monotonically to 69.35\% by iteration 5, expanding the supervision set to roughly 90\% of the training pixels through self-confidence above 0.95, plateauing 7.43 points below iSAGE. Varying the pseudo-label confidence threshold between 0.90 and 0.99 spans the iteration-5 mIoU from 69.00\% to 69.34\%, a 0.35 pp range that places the structural limit beyond threshold calibration. CRF-based label propagation peaks at 68.78\% by iteration 2 and then degenerates to 62.25\% by iteration 5, ending below every other baseline as spatial smoothing of softmax outputs compounds confident errors across iterations. iSAGE matches dense supervision while consuming eight thousand times fewer labeled pixels. Across four mechanisms, three oracle budget scales, and three pseudo-label confidence thresholds, no acquisition or expansion strategy reading the model's predictive distribution reaches iSAGE: the limit is not a budget question, a threshold question, or a smoothing question, it is a question of what information that distribution contains.

\subsection{Findings}\label{sec:disc-findings}

\subsubsection{Match-with-dense behavior}

The relationship between iSAGE and dense supervision depends on the quality of the dense ground truth, and the two datasets expose opposite ends of that dependence. BsB Aerial was annotated under controlled conditions by a single annotator with consistent boundary conventions, and iSAGE recovers 97.2\% of dense (74.79\% vs 76.93\%, 0.040\% labels). ISPRS Vaihingen carries known boundary inconsistencies across tiles, and iSAGE matches dense under identical protocol within 0.13 points (76.78\% vs 76.65\%). Dense supervision absorbs every inconsistency in the training labels and reproduces them at inference, while iSAGE labels only pixels where the class is visually unambiguous and is therefore less sensitive to boundary-label noise. The asymmetry is consistent with the framework's design rather than a chance effect: iSAGE matches dense when dense is clean and stays at parity when dense carries label noise. With two datasets at opposite ends of dense-label quality, this is a two-point observation rather than an established property. The reported numbers are also a lower bound: the one-pixel-per-class regime is the adversarial worst case, not a recommended budget, and practical deployments relax the cap. The match is additionally robust to the final-epoch versus best-validation reporting choice, with the gap between the two on iSAGE within 0.03 mIoU (Section~\ref{sec:vaihingen-results}).

This compression has a simple explanation. In a converged model most pixels are confidently correct, and a label on such a pixel contributes almost no gradient, since it only confirms what the model already predicts. The learning signal concentrates on the minority of pixels the model gets wrong, most of which lie at object boundaries or on small and high-variability classes. Dense supervision spends the overwhelming majority of its labels on the confident-correct majority, where they are redundant. iSAGE matches it not by extracting more from each label but by spending labels only where they carry information. The roughly eight-thousand-fold reduction is therefore expected rather than surprising, as it reflects the fraction of dense supervision that was ever informative.

\subsubsection{Per-class dynamics}

\begin{figure}[!htbp]
    \centering
    \includegraphics[width=0.9\columnwidth]{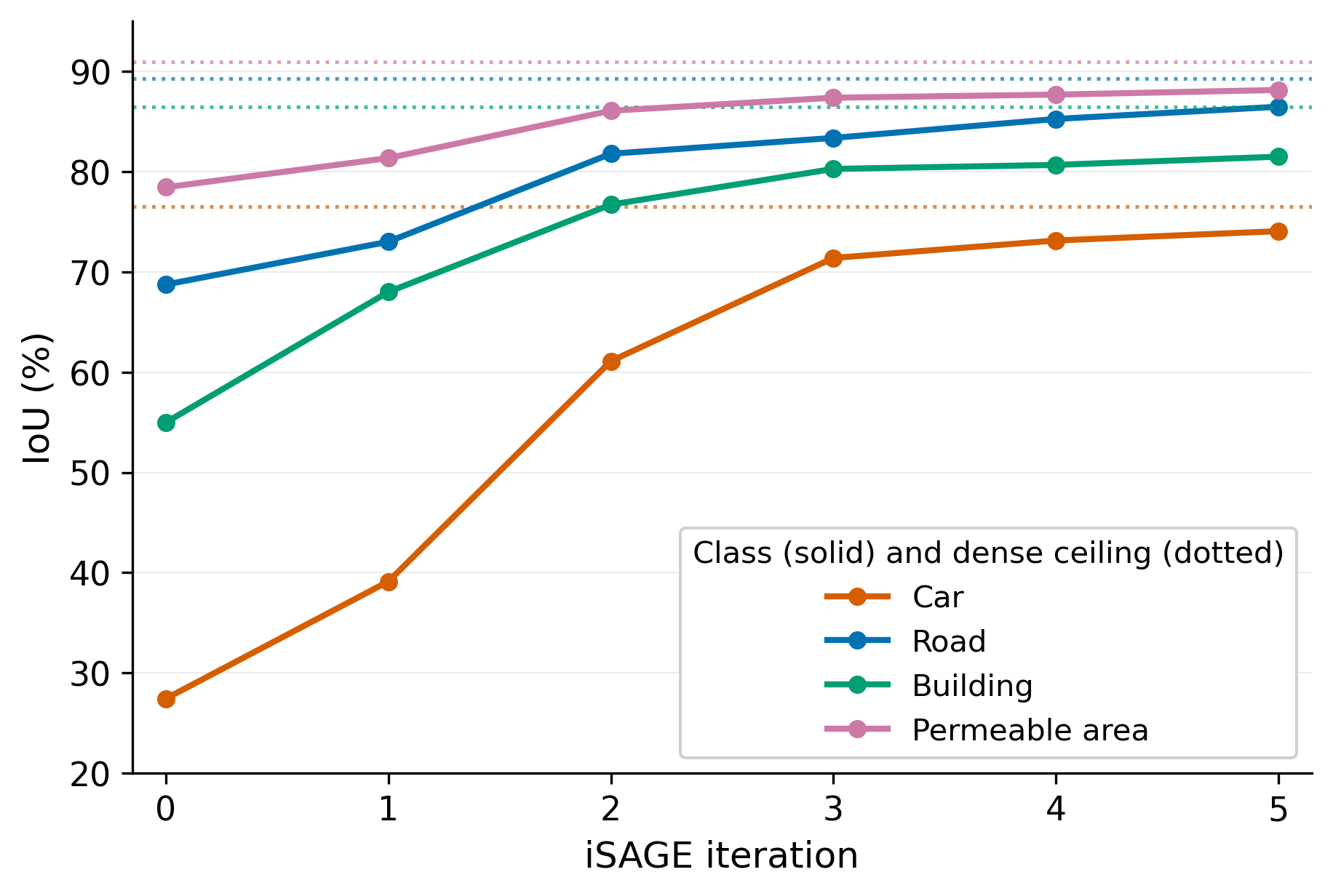}
    \caption{Per-class IoU convergence across iSAGE iterations on the BsB Aerial binary tasks (Table~\ref{tab:tab1}). Solid lines are iSAGE (EWDL) at each iteration; dotted lines mark the dense-supervision ceiling for each class. The stuff-versus-things contrast is visible: amorphous permeable areas start near the ceiling and gain little (+9.7 IoU over five iterations), whereas small cars start far below and climb steeply (+46.6 IoU), with roads and buildings in between.}
    \label{fig:fig7}
\end{figure}

The per-class pattern maps onto the stuff-versus-things distinction. iSAGE and random selection both recover most of dense performance on permeable areas (88.72\% iSAGE, 85.36\% random, 90.73\% dense, Table~\ref{tab:tab2}), but diverge sharply on cars (70.72\% iSAGE, 61.73\% random, 73.14\% dense). Targeted clicks reach the same neighborhood as dense supervision on cars (within 3 points) while random sampling stays 11 points behind. The binary experiments show the same pattern in isolation: permeable saturates from the seed, buildings and roads converge in a few iterations, cars require the most late-iteration effort. The iter-0$\to$iter-5 growth is +46.6 IoU on cars against +9.7 on permeable, with roads (+17.7) and buildings (+26.5) in between (Table~\ref{tab:tab1}, Figure~\ref{fig:fig7}). Two object properties explain the spread: small spatial extent dilutes the per-click gradient signal, and high shape variability surfaces new confident errors across iterations. These dynamics suggest a heuristic for budget planning: deprioritize homogeneous classes once they reach target performance, and reserve the late-iteration budget for small high-variability classes where confident errors persist.

\subsubsection{Architecture independence}

The match with dense is a property of the error-driven loop rather than of a particular backbone. Run unchanged across four architectures, iSAGE recovers between 93.2\% and 94.4\% of each architecture's own dense ceiling: U-Net with EfficientNet-B7 (94.4\%) and with ResNet-101 (93.4\%), DeepLabV3+ with ResNet-50 (93.2\%), and SegFormer with MiT-B2 (93.3\%) (Table~\ref{tab:arch}). The relative gap to dense is stable to about one point across both convolutional and transformer encoders, so the framework transfers to whatever segmentation model a deployment already uses.

\subsubsection{The loss helps slightly, the strategy drives the result}

The error-weighted loss gives a small benefit, but the real driver of performance is the error-driven strategy rather than the loss. Emphasizing hard pixels helps a little: trained from scratch on the same sparse annotation set, EWDL ($\lambda=5$, 74.79\%) and Focal (74.66\%) lead plain Dice ($\lambda=1$, 73.37\%) by about 1.4 points and match each other within standard deviation, which is expected because Focal applies a similar hard-example emphasis (Table~\ref{tab:tab2}, bottom panel). Over-amplification hurts ($\lambda=20$ falls to 72.18\%, Table~\ref{tab:ablation}). That margin is small next to the gap the strategy itself opens: under the same protocol, error-driven selection reaches 74.79\% while random selection at matched budget plateaus about 5 points lower (69.60\%). EWDL's value to iSAGE is therefore coherence rather than magnitude. The same indicator that defines where the annotator clicks ($\arg\max y_{pr}(x) \neq y_{gt}(x)$) defines where the loss amplifies the gradient, so every penalty pixel maps back to a specific click in the JSON record and the optimization step inherits the auditability of the acquisition step.

\subsection{Operational and cognitive properties of the workflow}\label{sec:disc-beyond-benchmark}

The properties discussed in this section are design consequences expected from the framework's architecture rather than measured outcomes, and quantifying them under realistic working conditions is left for future work.

\subsubsection{Incremental deployment and maintenance}

iSAGE's minimalism enables four operational capabilities that dense-mask pipelines do not provide by default. Deploying the model to fresh imagery reduces to opening the platform on the current model and clicking visible errors, with the existing annotation record remaining valid. Adding a new class costs only clicks on instances of that class, with no revisit to prior data. Correcting an annotation error is a single JSON edit, locatable by image and iteration without redrawing any region. Any past iteration is reloadable from its session directory, so state can be rewound, re-evaluated, or compared against alternative methods using the same record. These capabilities share one root: the annotation record is the dataset, the platform reads it directly, and no compiled mask artifact sits between human decision and gradient. In production contexts where models need ongoing maintenance (new sensors, monitored areas expanding, drift, evolving class priorities), dense-mask pipelines re-annotate from scratch at each increment, whereas iSAGE opens the platform on the current model and trains the next iteration at the cost of the clicks required. The cost scales with the change, not with the dataset.

\subsubsection{An auditable, versionable record}

The annotation record itself is a practical artifact. It is a small, human-readable list of tuples rather than a set of mask files, so it can be versioned and inspected like source code. The difference between two iterations is a handful of added clicks, and every label traces back to a specific human decision rather than sitting anonymously in a dense mask, which makes the supervision auditable and easy to share or reproduce.

\subsubsection{Directed rather than exhaustive attention}

The cognitive structure of the workflow mirrors this design. Dense polygon annotation requires sustained exhaustive attention: every pixel needs a label, including ambiguous boundary pixels where the annotator must decide without clear visual evidence. Fatigue degrades quality and speed, and the resulting errors are silent: a misplaced boundary is indistinguishable from a correct one in the final mask. iSAGE replaces exhaustive attention with directed attention: the annotator clicks only on pixels where visual evidence is unambiguous, skips the rest, and the training pass between iterations imposes a natural break. Deferral is free: a region the annotator cannot confidently classify at one iteration can wait for the next, where the model's updated prediction either resolves the ambiguity or surfaces it as an explicit error. The model and annotator function as mutual guides across iterations. The loop is also self-balancing: each click counts equally regardless of object size, and correcting a false positive shifts the decision boundary so the next iteration surfaces previously hidden false negatives, rebalancing the error distribution across iterations without explicit intervention.

\subsubsection{A model that is always usable and never final}

Because each iteration only adds clicks, the model is never final, and the practical stopping point is a visual judgment. The annotator watches the prediction overlay and the diminishing returns between rounds (on BsB Aerial multiclass, +9.51 mIoU from iteration 1 to 2 against +0.65 from iteration 5 to 6) and stops when the remaining errors are not worth correcting. The same loop makes recalibration cheap to check: running the current model on a new area, sensor, or acquisition surfaces its errors directly in the overlay, so confirming whether the model still holds, and adding a few clicks where it does not, costs only the clicks required.

\subsection{Limitations and Future Work}\label{sec:limitations}

iSAGE's empirical validation is confined to the aerial remote sensing family, matching the prevailing convention in human-in-the-loop semantic segmentation, where each framework is validated within a single imaging-domain family (aerial RS~\citep{lenczner2022dial, yang2024easyseg, hua2021semantic}, driving scenes~\citep{xie2022ripu, wu2022d2ada, guan2023ilmassl}, point clouds~\citep{liu2023ococ, siddiqui2020viewal}, or natural images~\citep{Bearman2016, lin2016scribblesup}). Within that family, the evidence already spans two sensors, two resolutions, two geographies, seven class types, and four architectures. The two claims that organize this paper have different reach: the structural argument that no function over the model's predictive distribution can distinguish confident errors from confident correct predictions holds wherever a single model reads its own outputs, whereas the effectiveness of the iSAGE framework is an empirical result established within this validated scope. The protocol carries no domain-specific priors: a click is an $(x, y, \mathrm{class})$ tuple, the JSON record is a flat list of such tuples, and EWDL operates on per-pixel correctness regardless of spectral or geometric properties. Extension to other sensors (SAR, multispectral, hyperspectral), modalities (medical, industrial inspection), or tasks (change detection, instance segmentation) therefore requires preprocessing, encoder, or UI adjustments rather than methodological changes. Automated acquisition and human correction target different supervision signals rather than competing: acquisition operates over predictive-distribution statistics, where confident errors are indistinguishable from confident correct predictions, while human correction supplies the externally-derived signal that distinguishes them, so combining automated coverage with human-targeted confident-error correction remains an open hybrid direction.

Several constraints frame the present design, each pointing to a clear evolution. The experiments use small patches ($256 \times 256$ on BsB Aerial and $512 \times 512$ on Vaihingen), where some objects are hard to label because they do not fit fully within the field of view. Larger patches would help on two fronts: each object's full extent becomes visible, making completeness easier to judge, and the annotation effort per total pixel drops further. Running iSAGE at larger patch sizes mainly requires a segmentation backbone that accommodates the larger input, a straightforward extension rather than a redesign. The annotation primitive is a single-pixel click, chosen for per-decision auditability. Scribbles, polygons, mixed-mode primitives, and exclusion clicks (``this pixel is not class $c$'') are additive extensions rather than replacements, and map naturally onto complementary-label losses studied for noisy-label learning. EWDL was designed for sparse supervision and may penalize pixel-level errors disproportionately under dense annotation with imperfect masks, but this case lies outside the protocol's premise. The workflow assumes an expert in the loop, which scopes its applicability to domains where class membership depends on conventions a non-expert cannot apply consistently, such as aerial imagery, medical segmentation, and industrial inspection.

Two further directions follow from the same record-as-dataset and integrated-loop infrastructure introduced here. Per-class behavior studies would map class characteristics (geometry, texture, instance scale, boundary type) to convergence behavior, turning the stuff-versus-things contrast observed here into a predictive model. Annotation-confidence characterization would directly measure where human clicks land in the model's predictive-distribution space, verifying that human correction concentrates in the confident-error region that the structural argument identifies as indistinguishable from confident correct predictions.

\FloatBarrier

\section{Conclusions}\label{sec:conclusion}
This work addressed a structural gap in sparse-supervised semantic segmentation: prior interactive frameworks combine human input with auxiliary machinery (pseudo-labels, propagation heuristics, uncertainty-based acquisition, consistency regularization, foundation-model labeling, domain adaptation) that operates on the model's predictive distribution, in which a confidently wrong pixel is indistinguishable from a confidently correct one by construction. iSAGE was proposed to investigate whether such machinery is necessary, configuring a framework in which the specialist's clicks on confident model errors constitute both the annotation record and the training supervision, the Error-Weighted Dice Loss amplifies the gradient at those labels, and an integrated software platform hosts inspection, annotation, and retraining as a continuous workflow.

Empirical validation on two aerial datasets shows that this minimalist configuration suffices to match dense fully-supervised performance under adversarial labeling regimes. iSAGE recovered 97.2\% of dense supervision on BsB Aerial (74.79\% mIoU at 0.040\% labeled pixels) and matched the dense baseline within 0.13 points on ISPRS Vaihingen (76.78\% vs 76.65\%), exceeding the published weakly-supervised methods evaluated on this benchmark. Four output-reading baselines plateaued below iSAGE under the same training protocol: oracle entropy with ground-truth labels at every query at 66.38\% mIoU (66.26 to 67.85\% across a 1 to 100$\times$ budget sweep), uniform random sampling within seed variance of the oracle, self-training pseudo-labels at 69.00 to 69.35\% across confidence thresholds 0.90 to 0.99, and CRF-based label propagation peaking at 68.78\% before degenerating to 62.25\% by iteration 5. These results are consistent with the gap being informational, neither a budget question, a confidence-threshold question, nor a smoothing question.

Across the 31-method landscape surveyed, iSAGE is the only iterative human-in-the-loop approach operating without auxiliary machinery, supporting the central claim that the accumulated machinery of prior pipelines is unnecessary when the human signal is delivered directly at confident errors. The open-source platform that enabled this validation also serves as a research instrument for follow-up work on cross-domain validation, annotation primitive comparisons, per-class convergence modeling, and inter-annotator dynamics within and beyond the validated remote sensing aerial scope.

% --- Back matter ---
\ifAnonymous
\else
\section*{CRediT authorship contribution statement}
\textbf{Osmar Luiz Ferreira de Carvalho:} Conceptualization, Methodology, Software, Validation, Formal analysis, Investigation, Data curation, Writing -- original draft, Visualization. \textbf{Osmar Ab\'ilio de Carvalho J\'unior:} Conceptualization, Validation, Resources, Writing -- review \& editing, Supervision, Project administration, Funding acquisition. \textbf{Anesmar Olino de Albuquerque:} Validation, Data curation, Writing -- review \& editing. \textbf{Daniel Guerreiro e Silva:} Conceptualization, Validation, Formal analysis, Investigation, Resources, Writing -- review \& editing, Supervision.
\fi

\section*{Disclosure statement}
The authors report there are no competing interests to declare.

\section*{Declaration of generative AI use}
% REQUIRED by GIScience & Remote Sensing. Confirm this reflects reality before
% submitting: if generative AI WAS used in the research or in writing the
% manuscript, disclose whether and how. Otherwise the suggested wording is below.
During the preparation of this manuscript, the authors used Claude (Anthropic) to assist with language editing and to improve the clarity and structure of the text. The authors reviewed and verified all content and take full responsibility for the publication.

\section*{Data availability statement}
\ifAnonymous
The iSAGE framework and code are available at \url{https://github.com/anonymized/iSAGE}, with the v1.0.0 release permanently archived at \url{https://doi.org/anonymized}. The BsB Aerial dataset and the session files for the experiments reported in this paper are available from the corresponding author upon reasonable request. The ISPRS Vaihingen dataset is available from the ISPRS 2D Semantic Labeling Contest benchmark at \url{https://www.isprs.org/resources/datasets/benchmarks/UrbanSemLab/2d-sem-label-vaihingen.aspx}.
\else
The data and code supporting the findings of this study are openly available. The iSAGE framework and code are available at \url{https://github.com/osmarluiz/iSAGE}, with the v1.0.0 release permanently archived on Zenodo (DOI: \url{https://doi.org/10.5281/zenodo.20596185}; \citealp{isage_zenodo}). The BsB Aerial dataset is openly available on Zenodo (DOI: \url{https://doi.org/10.5281/zenodo.20635237}; \citealp{bsbaerial_zenodo}), on Hugging Face (\url{https://huggingface.co/datasets/osmarluiz/BSB-Aerial-Dataset}), and at \url{https://github.com/osmarluiz/BSB-Aerial-Dataset}; the experiments reported here use a subset of this dataset. The ISPRS Vaihingen dataset is available from the ISPRS 2D Semantic Labeling Contest benchmark at \url{https://www.isprs.org/resources/datasets/benchmarks/UrbanSemLab/2d-sem-label-vaihingen.aspx}. The per-experiment session files (annotation records, masks, and trained models) are available from the corresponding author upon reasonable request.
\fi

\ifAnonymous
\else
\section*{Acknowledgements}
The authors thank the Laborat\'orio de Sistemas de Informa\c{c}\~oes Espaciais (LSIE) for providing the equipment and infrastructure necessary to carry out this research. This work was supported by the Coordena\c{c}\~ao de Aperfei\c{c}oamento de Pessoal de N\'ivel Superior (under grant 001), the Conselho Nacional de Desenvolvimento Cient\'ifico e Tecnol\'ogico (under grants 434838/2018-7 and 305769/2017-0), and by the DPI/BCE/UnB (under Edital n\textdegree{} 001/2025 DPI/BCE/UnB).
\fi

\section*{Funding}
This work was supported by the Coordena\c{c}\~ao de Aperfei\c{c}oamento de Pessoal de N\'ivel Superior (CAPES) under grant 001; the Conselho Nacional de Desenvolvimento Cient\'ifico e Tecnol\'ogico (CNPq) under grants 434838/2018-7 and 305769/2017-0; and the DPI/BCE/UnB under Edital n\textdegree{} 001/2025.

\ifAnonymous
\else
\section*{Notes on contributors}

\textit{Osmar Luiz Ferreira de Carvalho} is a PhD candidate in the Department of Electrical Engineering at the University of Bras\'ilia, Brazil. His research interests include computer vision, artificial intelligence, and remote sensing.

\textit{Osmar Ab\'ilio de Carvalho J\'unior} is a full professor in the Department of Geography at the University of Bras\'ilia, Brazil. His research interests include remote sensing and computer vision.

\textit{Anesmar Olino de Albuquerque} holds a PhD in Geography from the University of Bras\'ilia, Brazil. His research interests include remote sensing.

\textit{Daniel Guerreiro e Silva} is an adjunct professor in the Department of Electrical Engineering at the University of Bras\'ilia, Brazil. His research interests include computational intelligence, machine learning, and information theory.

\section*{ORCID}
\textit{Osmar Luiz Ferreira de Carvalho} \url{https://orcid.org/0000-0002-5619-8525}\\
\textit{Osmar Ab\'ilio de Carvalho J\'unior} \url{https://orcid.org/0000-0002-0346-1684}\\
\textit{Anesmar Olino de Albuquerque} \url{https://orcid.org/0000-0003-1561-7583}\\
\textit{Daniel Guerreiro e Silva} \url{https://orcid.org/0000-0003-2858-1110}
\fi

\bibliographystyle{tfcad}
\bibliography{bibliography}

@incollection{Lin2014Microsoft,
  booktitle={Computer Vision – ECCV 2014. Lecture Notes in Computer Science, vol 8693},
  doi={10.1007/978-3-319-10602-1_48},
  isbn={978-3-319-10601-4},
  issn=16113349,
  number={June},
  pmid=16190471,
  publisher={Springer Cham},
  address={Zurich, Switzerland},
  title={Microsoft COCO: Common Objects in Context},
  author={Lin, Tsung-Yi and Maire, Michael and Belongie, Serge and Hays, James and Perona, Pietro and Ramanan, Deva and Dollár, Piotr and Zitnick, C. Lawrence},
  editor={Fleet, David and Tomas, Pajdla and Schiele, Bernt and Tuytelaars, Tinne},
  pages={740--755},
  date=2014,
  year=2014,
  url={https://doi.org/10.1007/978-3-319-10602-1_48}
}

@inproceedings{cordts2016cityscapes,
  title={The cityscapes dataset for semantic urban scene understanding},
  author={Cordts, Marius and Omran, Mohamed and Ramos, Sebastian and Rehfeld, Timo and Enzweiler, Markus and Benenson, Rodrigo and Franke, Uwe and Roth, Stefan and Schiele, Bernt},
  booktitle={IEEE Conference on Computer Vision and Pattern Recognition (CVPR)},
  pages={3213--3223},
  doi={10.1109/CVPR.2016.350},
  url={https://doi.org/10.1109/CVPR.2016.350},
  year={2016}
}

@INPROCEEDINGS{deng2009imagenet,
  author={Deng, Jia and Dong, Wei and Socher, Richard and Li, Li-Jia and Kai Li and Li Fei-Fei},
  booktitle={2009 IEEE Conference on Computer Vision and Pattern Recognition}, 
  title={ImageNet: A large-scale hierarchical image database}, 
  year={2009},
  volume={},
  number={},
  pages={248-255},
  doi={10.1109/CVPR.2009.5206848},
  url={https://doi.org/10.1167/9.8.1037}}

@article{whang2023data,
  title={Data collection and quality challenges in deep learning: A data-centric AI perspective},
  author={Whang, Steven Euijong and Roh, Yuji and Song, Hwanjun and Lee, Jae-Gil},
  journal={The VLDB Journal},
  pages={1--23},
  year={2023},
  doi={10.1007/s00778-022-00775-9},
  url={https://doi.org/10.1007/s00778-022-00775-9},
  publisher={Springer}
}

@article{zha2023data,
  title={Data-centric artificial intelligence: A survey},
  author={Zha, Daochen and Bhat, Zaid Pervaiz and Lai, Kwei-Herng and Yang, Fan and Jiang, Zhimeng and Zhong, Shaochen and Hu, Xia},
  journal={arXiv preprint arXiv:2303.10158},
  eprint={2303.10158},
  archivePrefix={arXiv},
  primaryClass={cs.LG},
  url={https://arxiv.org/abs/2303.10158},
  year={2023}
}

@INPROCEEDINGS{sun2017revisiting,
  author={Sun, Chen and Shrivastava, Abhinav and Singh, Saurabh and Gupta, Abhinav},
  booktitle={2017 IEEE International Conference on Computer Vision (ICCV)}, 
  title={Revisiting Unreasonable Effectiveness of Data in Deep Learning Era}, 
  year={2017},
  volume={},
  number={},
  pages={843-852},
  doi={10.1109/ICCV.2017.97},
  url={https://doi.org/10.1109/ICCV.2017.97}}

@article{Rottensteiner2014,
author = {Rottensteiner, Franz and Sohn, Gunho and Gerke, Markus and Wegner, Jan Dirk and Breitkopf, Uwe and Jung, Jaewook},
doi = {10.1016/j.isprsjprs.2013.10.004},
issn = {09242716},
journal = {ISPRS Journal of Photogrammetry and Remote Sensing},
pages = {256--271},
publisher = {International Society for Photogrammetry and Remote Sensing, Inc. (ISPRS)},
title = {{Results of the ISPRS benchmark on urban object detection and 3D building reconstruction}},
url = {https://doi.org/10.1016/j.isprsjprs.2013.10.004},
volume = {93},
year = {2014}
}

@misc{Wang2021,
      title={LoveDA: A Remote Sensing Land-Cover Dataset for Domain Adaptive Semantic Segmentation}, 
      author={Junjue Wang and Zhuo Zheng and Ailong Ma and Xiaoyan Lu and Yanfei Zhong},
      year={2022},
      eprint={2110.08733},
      archivePrefix={arXiv},
      primaryClass={cs.CV},
      url={https://doi.org/10.48550/arXiv.2110.08733}, 
}

@article{de2022panoptic,
  title={Panoptic Segmentation Meets Remote Sensing},
  author={de Carvalho, Osmar Luiz Ferreira and de Carvalho J{\'u}nior, Osmar Ab{\'\i}lio and Silva, Cristiano Rosa e and de Albuquerque, Anesmar Olino and Santana, Nickolas Castro and Borges, Dibio Leandro and Gomes, Roberto Arnaldo Trancoso and Guimar{\~a}es, Renato Fontes},
  journal={Remote Sensing},
  volume={14},
  number={4},
  pages={965},
  year={2022},
  doi= {10.3390/rs14040965},
  publisher={MDPI},
  url={https://doi.org/10.3390/rs14040965}
}

@article{Rahnemoonfar2023,
author = {Rahnemoonfar, Maryam and Chowdhury, Tashnim and Murphy, Robin},
doi = {10.1038/s41597-023-02799-4},
issn = {2052-4463},
journal = {Scientific Data},
month = {dec},
number = {1},
pages = {913},
title = {{RescueNet: A High Resolution UAV Semantic Segmentation Dataset for Natural Disaster Damage Assessment}},
url = {https://doi.org/10.1038/s41597-023-02799-4},
volume = {10},
year = {2023}
}

@article{Bearman2016,
archivePrefix = {arXiv},
arxivId = {1506.02106},
author = {Bearman, Amy and Russakovsky, Olga and Ferrari, Vittorio and Fei-Fei, Li},
doi = {10.1007/978-3-319-46478-7_34},
eprint = {1506.02106},
isbn = {9783319464770},
issn = {16113349},
journal = {Lecture Notes in Computer Science (including subseries Lecture Notes in Artificial Intelligence and Lecture Notes in Bioinformatics)},
pages = {549--565},
title = {{What's the point: Semantic segmentation with point supervision}},
volume = {9911 LNCS},
year = {2016},
url={https://doi.org/10.1007/978-3-319-46478-7_34}
}

@article{hua2021semantic,
  title={Semantic segmentation of remote sensing images with sparse annotations},
  author={Hua, Yuansheng and Marcos, Diego and Mou, Lichao and Zhu, Xiao Xiang and Tuia, Devis},
  journal={IEEE Geoscience and Remote Sensing Letters},
  volume={19},
  pages={1-5},
  year={2021},
  doi={10.1109/LGRS.2021.3051053},
  publisher={IEEE},
  url={https://doi.org/10.1109/LGRS.2021.3051053}
}

@inproceedings{sener2018active,
  title={Active learning for convolutional neural networks: A core-set approach},
  author={Sener, Ozan and Savarese, Silvio},
  booktitle={International Conference on Learning Representations (ICLR)},
  year={2018},
  eprint={1708.00489},
  archivePrefix={arXiv},
  primaryClass={cs.LG},
  url={https://arxiv.org/abs/1708.00489}
}

@inproceedings{lee2013pseudo,
  title={Pseudo-label: The simple and efficient semi-supervised learning method for deep neural networks},
  author={Lee, Dong-Hyun and others},
  booktitle={Workshop on challenges in representation learning, ICML},
  volume={3},
  number={2},
  pages={896},
  year={2013},
  organization={Atlanta}
}

@inproceedings{kirillov2023sam,
  author={Kirillov, Alexander and Mintun, Eric and Ravi, Nikhila and Mao, Hanzi and Rolland, Chloe and Gustafson, Laura and Xiao, Tete and Whitehead, Spencer and Berg, Alexander C. and Lo, Wan-Yen and Dollár, Piotr and Girshick, Ross},
  booktitle={2023 IEEE/CVF International Conference on Computer Vision (ICCV)}, 
  title={Segment Anything}, 
  year={2023},
  volume={},
  number={},
  pages={3992-4003},
  url={https://doi.org/10.1109/ICCV51070.2023.00371},
  doi={10.1109/ICCV51070.2023.00371}}

@inproceedings{lin2016scribblesup,
  author={Lin, Di and Dai, Jifeng and Jia, Jiaya and He, Kaiming and Sun, Jian},
  booktitle={2016 IEEE Conference on Computer Vision and Pattern Recognition (CVPR)}, 
  title={ScribbleSup: Scribble-Supervised Convolutional Networks for Semantic Segmentation}, 
  year={2016},
  volume={},
  number={},
  pages={3159-3167},
  url = {https://doi.org/10.1109/CVPR.2016.344},
  doi={10.1109/CVPR.2016.344}}

@incollection{Cicek2016,
address = {Cham},
author = {{\c{C}}i{\c{c}}ek, {\"{O}}zg{\"{u}}n and Abdulkadir, Ahmed and Lienkamp, Soeren S. and Brox, Thomas and Ronneberger, Olaf},
booktitle = {Medical Image Computing and Computer-Assisted Intervention – MICCAI 2016},
doi = {10.1007/978-3-319-46723-8_49},
editor = {{Ourselin, S.} and Joskowicz, L. and Sabuncu, M. and Unal, G. and Wells, W.},
pages = {424--432},
publisher = {Springer},
title = {{3D U-Net: Learning Dense Volumetric Segmentation from Sparse Annotation}},
url = {https://doi.org/10.1007/978-3-319-46723-8_49},
year = {2016}
}

@article{gao2022segmentation,
   author = {Feng Gao and Minhao Hu and Min Er Zhong and Shixiang Feng and Xuwei Tian and Xiaochun Meng and Ma yi di li Ni-jia-ti and Zeping Huang and Minyi Lv and Tao Song and Xiaofan Zhang and Xiaoguang Zou and Xiaojian Wu},
   doi = {10.1016/j.media.2022.102515},
   issn = {13618423},
   journal = {Medical Image Analysis},
   month = {8},
   pmid = {35780593},
   publisher = {Elsevier B.V.},
   title = {Segmentation only uses sparse annotations: Unified weakly and semi-supervised learning in medical images},
   volume = {80},
   year = {2022},
   url = {https://doi.org/10.1016/j.media.2022.102515}}

@article{Kervadec2019,
author = {Kervadec, Hoel and Dolz, Jose and Tang, Meng and Granger, Eric and Boykov, Yuri and {Ben Ayed}, Ismail},
doi = {10.1016/j.media.2019.02.009},
issn = {13618415},
journal = {Medical Image Analysis},
month = {may},
pages = {88--99},
title = {{Constrained-CNN losses for weakly supervised segmentation}},
url = {https://doi.org/10.1016/10.1016/j.media.2019.02.009},
volume = {54},
year = {2019}
}

@article{Yang2020,
author = {Yang, Guanyu and Wang, Chuanxia and Yang, Jian and Chen, Yang and Tang, Lijun and Shao, Pengfei and Dillenseger, Jean-Louis and Shu, Huazhong and Luo, Limin},
doi = {10.1186/s12880-020-00435-w},
issn = {1471-2342},
journal = {BMC Medical Imaging},
month = {dec},
number = {1},
pages = {37},
title = {{Weakly-supervised convolutional neural networks of renal tumor segmentation in abdominal CTA images}},
url = {https://doi.org/10.1186/s12880-020-00435-w},
volume = {20},
year = {2020}
}

@article{Liu2023,
author = {Liu, Xiaoming and Liu, Qi and Zhang, Ying and Wang, Man and Tang, Jinshan},
doi = {10.1016/j.compbiomed.2022.106467},
issn = {00104825},
journal = {Computers in Biology and Medicine},
month = {feb},
pages = {106467},
title = {{TSSK-Net: Weakly supervised biomarker localization and segmentation with image-level annotation in retinal OCT images}},
url = {https://doi.org/10.1016/j.compbiomed.2022.106467},
volume = {153},
year = {2023}
}

@article{Maggiolo2022,
author = {Maggiolo, Luca and Marcos, Diego and Moser, Gabriele and Serpico, Sebastiano B. and Tuia, Devis},
doi = {10.1109/TGRS.2021.3095832},
issn = {15580644},
journal = {IEEE Transactions on Geoscience and Remote Sensing},
pages = {1-15},
publisher = {IEEE},
title = {{A Semisupervised CRF Model for CNN-Based Semantic Segmentation with Sparse Ground Truth}},
volume = {60},
year = {2022},
url = {https://doi.org/10.1109/TGRS.2021.3095832}
}

@article{Mazhar2022,
author = {Mazhar, Sarah and Sun, Guangmin and Bilal, Anas and Hassan, Bilal and Li, Yu and Zhang, Junjie and Lin, Yinyi and Khan, Ali and Ahmed, Ramsha and Hassan, Taimur},
doi = {10.3390/rs14143283},
issn = {20724292},
journal = {Remote Sensing},
number = {14},
pages = {1-18},
title = {{AUnet: A Deep Learning Framework for Surface Water Channel Mapping Using Large-Coverage Remote Sensing Images and Sparse Scribble Annotations from OSM Data}},
volume = {14},
year = {2022},
url = {https://doi.org/10.3390/rs14143283}
}

@article{Gbodjo2021,
author = {Gbodjo, Yawogan Jean Eudes and Montet, Olivier and Ienco, Dino and Gaetano, Raffaele and Dupuy, Stephane},
doi = {10.1109/JSTARS.2021.3119191},
issn = {1939-1404},
journal = {IEEE Journal of Selected Topics in Applied Earth Observations and Remote Sensing},
pages = {11485--11499},
publisher = {IEEE},
title = {{Multisensor Land Cover Classification With Sparsely Annotated Data Based on Convolutional Neural Networks and Self-Distillation}},
url = {https://doi.org/10.1109/JSTARS.2021.3119191},
volume = {14},
year = {2021}
}

@inproceedings{liang2022tree,
   author = {Zhiyuan Liang and Tiancai Wang and Xiangyu Zhang and Jian Sun and Jianbing Shen},
   doi = {10.1109/CVPR52688.2022.01640},
   isbn = {978-1-6654-6946-3},
   booktitle = {2022 IEEE/CVF Conference on Computer Vision and Pattern Recognition (CVPR)},
   month = {6},
   pages = {16886-16895},
   publisher = {IEEE},
   title = {Tree Energy Loss: Towards Sparsely Annotated Semantic Segmentation},
   url = {https://doi.org/10.1109/CVPR52688.2022.01640},
   year = {2022}
}

@inproceedings{liu2021one,
   author = {Zhengzhe Liu and Xiaojuan Qi and Chi-Wing Fu},
   doi = {10.1109/CVPR46437.2021.00177},
   isbn = {978-1-6654-4509-2},
   booktitle = {2021 IEEE/CVF Conference on Computer Vision and Pattern Recognition (CVPR)},
   month = {6},
   pages = {1726-1736},
   publisher = {IEEE},
   title = {One Thing One Click: A Self-Training Approach for Weakly Supervised 3D Semantic Segmentation},
   url = {https://doi.org/10.1109/CVPR46437.2021.00177},
   year = {2021}
}

@misc{chen2017rethinking,
      title={Rethinking Atrous Convolution for Semantic Image Segmentation}, 
      author={Liang-Chieh Chen and George Papandreou and Florian Schroff and Hartwig Adam},
      year={2017},
      eprint={1706.05587},
      archivePrefix={arXiv},
      primaryClass={cs.CV},
      url={https://arxiv.org/abs/1706.05587}}

@incollection{can2018learning,
address = {Cham},
archivePrefix = {arXiv},
arxivId = {1807.04668},
author = {Can, Yigit B. and Chaitanya, Krishna and Mustafa, Basil and Koch, Lisa M. and Konukoglu, Ender and Baumgartner, Christian F.},
booktitle = {Lecture Notes in Computer Science (including subseries Lecture Notes in Artificial Intelligence and Lecture Notes in Bioinformatics)},
doi = {10.1007/978-3-030-00889-5_27},
editor = {Stoyanov, Danail and Taylor, Zeike and Carneiro, Gustavo and Syeda-Mahmood, Tanveer and Martel, Anne and Maier-Hein, Lena and Tavares, Jo{\~{a}}o Manuel R.S. and Bradley, Andrew and Papa, Jo{\~{a}}o Paulo and Belagiannis, Vasileios and Nascimento, Jacinto C. and Lu, Zhi and Conjeti, Sailesh and Moradi, Mehdi and Greenspan, Hayit and Madabhushi, Anant},
eprint = {1807.04668},
isbn = {9783030008888},
issn = {16113349},
pages = {236--244},
publisher = {Springer International Publishing},
title = {Learning to Segment Medical Images with Scribble-Supervision Alone},
url = {https://doi.org/10.1007/978-3-030-00889-5_27},
volume = {11045 LNCS},
year = {2018}
}

@article{arnab2018conditional,
author = {Arnab, Anurag and Zheng, Shuai and Jayasumana, Sadeep and Romera-Paredes, Bernardino and Larsson, Mans and Kirillov, Alexander and Savchynskyy, Bogdan and Rother, Carsten and Kahl, Fredrik and Torr, Philip H.S.},
doi = {10.1109/MSP.2017.2762355},
issn = {1053-5888},
journal = {IEEE Signal Processing Magazine},
month = {jan},
number = {1},
pages = {37--52},
title = {{Conditional Random Fields Meet Deep Neural Networks for Semantic Segmentation: Combining Probabilistic Graphical Models with Deep Learning for Structured Prediction}},
url = {https://doi.org/10.1109/MSP.2017.2762355},
volume = {35},
year = {2018}}

@article{wang2022pymic,
title = {PyMIC: A deep learning toolkit for annotation-efficient medical image segmentation},
journal = {Computer Methods and Programs in Biomedicine},
volume = {231},
pages = {107398},
year = {2023},
issn = {0169-2607},
doi = {10.1016/j.cmpb.2023.107398},
url = {https://doi.org/10.1016/j.cmpb.2023.107398},
author = {Guotai Wang and Xiangde Luo and Ran Gu and Shuojue Yang and Yijie Qu and Shuwei Zhai and Qianfei Zhao and Kang Li and Shaoting Zhang}}

@inproceedings{belharbi2021deep,
   author = {Soufiane Belharbi and Ismail Ben Ayed and Luke McCaffrey and Eric Granger},
   doi = {10.1109/WACV48630.2021.00338},
   isbn = {978-1-6654-0477-8},
   booktitle = {2021 IEEE Winter Conference on Applications of Computer Vision (WACV)},
   month = {1},
   pages = {3337-3346},
   publisher = {IEEE},
   title = {Deep Active Learning for Joint Classification \& Segmentation with Weak Annotator},
   url = {https://doi.org/10.1109/WACV48630.2021.00338},
   year = {2021}
}

@article{ren2022framework,
   author = {Quan Ren and Hongbing Zhang and Dailu Zhang and Xiang Zhao and Lizhi Yan and Jianwen Rui and Fanxin Zeng and Xinyi Zhu},
   doi = {10.1016/j.eswa.2022.117278},
   issn = {09574174},
   journal = {Expert Systems with Applications},
   month = {9},
   publisher = {Elsevier Ltd},
   title = {A framework of active learning and semi-supervised learning for lithology identification based on improved naive Bayes},
   volume = {202},
   year = {2022},
   url={https://doi.org/10.1016/j.eswa.2022.117278}
}

@inproceedings{desai2022active,
   author = {Shasvat Desai and Debasmita Ghose},
   doi = {10.1109/WACV51458.2022.00155},
   isbn = {978-1-6654-0915-5},
   booktitle = {2022 IEEE/CVF Winter Conference on Applications of Computer Vision (WACV)},
   month = {1},
   pages = {1485-1495},
   publisher = {IEEE},
   title = {Active Learning for Improved Semi-Supervised Semantic Segmentation in Satellite Images},
   url = {https://doi.org/10.1109/WACV51458.2022.00155},
   year = {2022}
}

@article{alonso2019coralseg,
author = {Alonso, I{\~{n}}igo and Yuval, Matan and Eyal, Gal and Treibitz, Tali and Murillo, Ana C.},
doi = {10.1002/rob.21915},
issn = {15564967},
journal = {Journal of Field Robotics},
number = {8},
pages = {1456-1477},
title = {{CoralSeg: Learning coral segmentation from sparse annotations}},
volume = {36},
year = {2019},
url={https://doi.org/10.1002/rob.21915}
}

@incollection{Lee2020,
address = {Cham},
archivePrefix = {arXiv},
arxivId = {2006.12890},
author = {Lee, Hyeonsoo and Jeong, Won-Ki},
booktitle = {Lecture Notes in Computer Science (including subseries Lecture Notes in Artificial Intelligence and Lecture Notes in Bioinformatics)},
doi = {10.1007/978-3-030-59710-8_2},
editor = {Martel, Anne L. and Abolmaesumi, Purang and Stoyanov, Danail and Mateus, Diana and Zuluaga, Maria A. and Zhou, S. Kevin and Racoceanu, Daniel and Joskowicz, Leo},
eprint = {2006.12890},
isbn = {9783030597092},
issn = {16113349},
pages = {14--23},
publisher = {Springer International Publishing},
title = {{Scribble2Label: Scribble-Supervised Cell Segmentation via Self-generating Pseudo-Labels with Consistency}},
url = {https://doi.org/10.1007/978-3-030-59710-8_2},
volume = {12261 LNCS},
year = {2020}}

@article{kellenberger2019half,
  title={Half a percent of labels is enough: Efficient animal detection in UAV imagery using deep CNNs and active learning},
  author={Kellenberger, Benjamin and Marcos, Diego and Lobry, Sylvain and Tuia, Devis},
  journal={IEEE Transactions on Geoscience and Remote Sensing},
  volume={57},
  number={12},
  pages={9524-9533},
  year={2019},
  doi={10.1109/TGRS.2019.2927393},
  url={https://doi.org/10.1109/TGRS.2019.2927393},
  publisher={IEEE}
}

@inproceedings{yoo2019learning,
   author = {Donggeun Yoo and In So Kweon},
   doi = {10.1109/CVPR.2019.00018},
   isbn = {978-1-7281-3293-8},
   booktitle = {2019 IEEE/CVF Conference on Computer Vision and Pattern Recognition (CVPR)},
   month = {6},
   pages = {93-102},
   publisher = {IEEE},
   title = {Learning Loss for Active Learning},
   url = {https://doi.org/10.1109/CVPR.2019.00018},
   year = {2019}
}

@inproceedings{lai2021joint,
   author = {Zhengfeng Lai and Chao Wang and Luca Cerny Oliveira and Brittany N. Dugger and Sen-Ching Cheung and Chen-Nee Chuah},
   doi = {10.1109/ICCVW54120.2021.00072},
   isbn = {978-1-6654-0191-3},
   booktitle = {2021 IEEE/CVF International Conference on Computer Vision Workshops (ICCVW)},
   month = {10},
   pages = {591-600},
   publisher = {IEEE},
   title = {Joint Semi-supervised and Active Learning for Segmentation of Gigapixel Pathology Images with Cost-Effective Labeling},
   url = {https://doi.org/10.1109/ICCVW54120.2021.00072},
   year = {2021}
}

@inproceedings{yuan2021multiple,
   author = {Tianning Yuan and Fang Wan and Mengying Fu and Jianzhuang Liu and Songcen Xu and Xiangyang Ji and Qixiang Ye},
   doi = {10.1109/CVPR46437.2021.00529},
   isbn = {978-1-6654-4509-2},
   booktitle = {2021 IEEE/CVF Conference on Computer Vision and Pattern Recognition (CVPR)},
   month = {6},
   pages = {5326-5335},
   publisher = {IEEE},
   title = {Multiple Instance Active Learning for Object Detection},
   url = {https://doi.org/10.1109/CVPR46437.2021.00529},
   year = {2021}
}

@article{fan2024integrating,
   author = {Cheng Fan and Qiuting Wu and Yang Zhao and Like Mo},
   doi = {10.1016/j.apenergy.2023.122356},
   issn = {03062619},
   journal = {Applied Energy},
   month = {2},
   publisher = {Elsevier Ltd},
   title = {Integrating active learning and semi-supervised learning for improved data-driven HVAC fault diagnosis performance},
   volume = {356},
   year = {2024},
   url={https://doi.org/10.1016/j.apenergy.2023.122356}
}

@inproceedings{chen2024think,
   author = {Jiayi Chen and Benteng Ma and Hengfei Cui and Yong Xia},
   doi = {10.1109/CVPR52733.2024.01087},
   isbn = {979-8-3503-5300-6},
   booktitle = {2024 IEEE/CVF Conference on Computer Vision and Pattern Recognition (CVPR)},
   month = {6},
   pages = {11439-11449},
   publisher = {IEEE},
   title = {Think Twice Before Selection: Federated Evidential Active Learning for Medical Image Analysis with Domain Shifts},
   url = {https://doi.org/10.1109/CVPR52733.2024.01087},
   year = {2024}
}

@inproceedings{yamani2024active,
   author = {Asma Yamani and Albandari Alyami and Hamzah Luqman and Bernard Ghanem and Silvio Giancola},
   doi = {10.1109/WACV57701.2024.00187},
   isbn = {979-8-3503-1892-0},
   booktitle = {2024 IEEE/CVF Winter Conference on Applications of Computer Vision (WACV)},
   month = {1},
   pages = {1849-1858},
   publisher = {IEEE},
   title = {Active Learning for Single-Stage Object Detection in UAV Images},
   url = {https://doi.org/10.1109/WACV57701.2024.00187},
   year = {2024}
}

@misc{ge2024esa,
      title={ESA: Annotation-Efficient Active Learning for Semantic Segmentation},
      author={Jinchao Ge and Zeyu Zhang and Minh Hieu Phan and Bowen Zhang and Akide Liu and Yang Zhao and Shuwen Zhao},
      year={2024},
      eprint={2408.13491},
      archivePrefix={arXiv},
      primaryClass={cs.CV},
      url={https://arxiv.org/abs/2408.13491}}

@article{yang2024easyseg,
   author = {Liangzhe Yang and Hao Chen and Anran Yang and Jun Li},
   doi = {10.1109/TGRS.2024.3399260},
   issn = {15580644},
   journal = {IEEE Transactions on Geoscience and Remote Sensing},
   pages = {1-18},
   publisher = {Institute of Electrical and Electronics Engineers Inc.},
   title = {EasySeg: An Error-Aware Domain Adaptation Framework for Remote Sensing Imagery Semantic Segmentation via Interactive Learning and Active Learning},
   volume = {62},
   year = {2024},
   url={https://doi.org/10.1109/TGRS.2024.3399260}
}

@article{de2022bounding,
  title={Bounding box-free instance segmentation using semi-supervised iterative learning for vehicle detection},
  author={de Carvalho, Osmar Luiz Ferreira and de Carvalho J{\'u}nior, Osmar Ab{\'\i}lio and de Albuquerque, Anesmar Olino and Santana, Nickolas Castro and Guimar{\~a}es, Renato Fontes and Gomes, Roberto Arnaldo Trancoso and Borges, D{\'\i}bio Leandro},
  journal={IEEE Journal of Selected Topics in Applied Earth Observations and Remote Sensing},
  volume={15},
  pages={3403--3420},
  year={2022},
  doi={10.1109/JSTARS.2022.3169128},
  publisher={IEEE},
  url={https://doi.org/10.1109/JSTARS.2022.3169128}
}

@article{de2022rethinking,
  author={de Carvalho, Osmar L. F. and de Carvalho Júnior, Osmar A. and de Albuquerque, Anesmar O. and Santana, Nickolas C. and Borges, Díbio L.},
  journal={IEEE Geoscience and Remote Sensing Letters}, 
  title={Rethinking Panoptic Segmentation in Remote Sensing: A Hybrid Approach Using Semantic Segmentation and Non-Learning Methods}, 
  year={2022},
  volume={19},
  number={},
  pages={1-5},
  doi={10.1109/LGRS.2022.3172207},
  url={https://doi.org/10.1109/LGRS.2022.3172207}}

@INPROCEEDINGS{wang2019panet,
  author={Wang, Kaixin and Liew, Jun Hao and Zou, Yingtian and Zhou, Daquan and Feng, Jiashi},
  booktitle={2019 IEEE/CVF International Conference on Computer Vision (ICCV)}, 
  title={PANet: Few-Shot Image Semantic Segmentation With Prototype Alignment}, 
  year={2019},
  volume={},
  number={},
  pages={9196-9205},
  doi={10.1109/ICCV.2019.00929},
  url={https://doi.org/10.1109/ICCV.2019.00929}}

@misc{boudiaf2021fewshot,
      title={Few-Shot Segmentation Without Meta-Learning: A Good Transductive Inference Is All You Need?}, 
      author={Malik Boudiaf and Hoel Kervadec and Ziko Imtiaz Masud and Pablo Piantanida and Ismail Ben Ayed and Jose Dolz},
      year={2021},
      eprint={2012.06166},
      archivePrefix={arXiv},
      primaryClass={cs.CV},
      url={https://arxiv.org/abs/2012.06166}, 
}

@INPROCEEDINGS{luddecke2022image,
  author={Lüddecke, Timo and Ecker, Alexander},
  booktitle={2022 IEEE/CVF Conference on Computer Vision and Pattern Recognition (CVPR)}, 
  title={Image Segmentation Using Text and Image Prompts}, 
  year={2022},
  volume={},
  number={},
  pages={7076-7086},
  doi={10.1109/CVPR52688.2022.00695},
  url={https://doi.org/10.1109/CVPR52688.2022.00695}}

@misc{zou2023segment,
      title={Segment Everything Everywhere All at Once}, 
      author={Xueyan Zou and Jianwei Yang and Hao Zhang and Feng Li and Linjie Li and Jianfeng Wang and Lijuan Wang and Jianfeng Gao and Yong Jae Lee},
      year={2023},
      eprint={2304.06718},
      archivePrefix={arXiv},
      primaryClass={cs.CV},
      url={https://arxiv.org/abs/2304.06718}, 
}

@misc{shaban2017one,
      title={One-Shot Learning for Semantic Segmentation}, 
      author={Amirreza Shaban and Shray Bansal and Zhen Liu and Irfan Essa and Byron Boots},
      year={2017},
      eprint={1709.03410},
      archivePrefix={arXiv},
      primaryClass={cs.CV},
      url={https://arxiv.org/abs/1709.03410}}

@article{li2023hal,
   author = {Xiaokang Li and Menghua Xia and Jing Jiao and Shichong Zhou and Cai Chang and Yuanyuan Wang and Yi Guo},
   doi = {10.1016/j.media.2023.102862},
   issn = {13618423},
   journal = {Medical Image Analysis},
   month = {8},
   pmid = {37295312},
   publisher = {Elsevier B.V.},
   title = {HAL-IA: A Hybrid Active Learning framework using Interactive Annotation for medical image segmentation},
   volume = {88},
   year = {2023},
   url={https://doi.org/10.1016/j.media.2023.102862}
}

@article{jin2022cold,
   author = {Qiuye Jin and Mingzhi Yuan and Shiman Li and Haoran Wang and Manning Wang and Zhijian Song},
   doi = {10.1016/j.ins.2022.10.066},
   issn = {00200255},
   journal = {Information Sciences},
   month = {11},
   pages = {16-36},
   publisher = {Elsevier Inc.},
   title = {Cold-start active learning for image classification},
   volume = {616},
   year = {2022},
   url={https://doi.org/10.1016/j.ins.2022.10.066}
}

@inproceedings{teng2022structured,
   author={Teng, Yao and Wang, Limin},
   doi = {10.1109/CVPR52688.2022.01883},
   isbn = {978-1-6654-6946-3},
   booktitle = {2022 IEEE/CVF Conference on Computer Vision and Pattern Recognition (CVPR)},
   month = {6},
   pages = {19415-19424},
   publisher = {IEEE},
   title = {Structured Sparse R-CNN for Direct Scene Graph Generation},
   year = {2022},
   url= {https://doi.org/10.1109/CVPR52688.2022.01883}
}

@inproceedings{rangnekar2023semantic,
   author = {Aneesh Rangnekar and Christopher Kanan and Matthew Hoffman},
   doi = {10.1109/WACV56688.2023.00591},
   isbn = {978-1-6654-9346-8},
   booktitle = {2023 IEEE/CVF Winter Conference on Applications of Computer Vision (WACV)},
   month = {1},
   pages = {5955-5966},
   publisher = {IEEE},
   title = {Semantic Segmentation with Active Semi-Supervised Learning},
   year = {2023},
   url={https://doi.org/10.1109/WACV56688.2023.00591}
}

@misc{iakubovskii2019,
  Author = {Pavel Iakubovskii},
  Title = {Segmentation Models Pytorch},
  Year = {2019},
  Publisher = {GitHub},
  Journal = {GitHub repository},
  Howpublished = {\url{https://github.com/qubvel/segmentation_models.pytorch}}
}

@inproceedings{ronneberger2015unet,
  title={U-net: Convolutional networks for biomedical image segmentation},
  author={Ronneberger, Olaf and Fischer, Philipp and Brox, Thomas},
  booktitle={International Conference on Medical Image Computing and Computer-assisted Intervention},
  pages={234--241},
  year={2015},
  doi={10.1007/978-3-319-24574-4_28},
  organization={Springer},
  url={https://doi.org/10.1007/978-3-319-24574-4_28}
}

@inproceedings{tan2019efficientnet,
  title = {{E}fficient{N}et: Rethinking Model Scaling for Convolutional Neural Networks},
  author = {Tan, Mingxing and Le, Quoc},
  booktitle = {Proceedings of the 36th International Conference on Machine Learning},
  pages = {6105--6114},
  year = 	{2019},
  editor = {Chaudhuri, Kamalika and Salakhutdinov, Ruslan},
  volume = {97},
  series = {Proceedings of Machine Learning Research},
  month = {09--15 Jun},
  eprint={1905.11946},
  archivePrefix={arXiv},
  primaryClass={cs.LG},
  publisher = {PMLR},
  url = {https://proceedings.mlr.press/v97/tan19a.html}
}

@INPROCEEDINGS{siddiqui2020viewal,
  author={Siddiqui, Yawar and Valentin, Julien and Niessner, Matthias},
  booktitle={2020 IEEE/CVF Conference on Computer Vision and Pattern Recognition (CVPR)}, 
  title={ViewAL: Active Learning With Viewpoint Entropy for Semantic Segmentation}, 
  year={2020},
  volume={},
  number={},
  pages={9430-9440},
  doi={10.1109/CVPR42600.2020.00945},
  url={https://doi.org/10.1109/CVPR42600.2020.00945}}

@article{lenczner2022dial,
  title={DIAL: Deep interactive and active learning for semantic segmentation in remote sensing},
  author={Lenczner, Gaston and Chan-Hon-Tong, Adrien and Le Saux, Bertrand and Luminari, Nicola and Le Besnerais, Guy},
  journal={IEEE Journal of Selected Topics in Applied Earth Observations and Remote Sensing},
  volume={15},
  pages={3376--3389},
  year={2022},
  doi={10.1109/JSTARS.2022.3166551},
  publisher={IEEE},
  url={https://doi.org/10.1109/JSTARS.2022.3166551}
}

@INPROCEEDINGS{khoreva2017simple,
  author={Khoreva, Anna and Benenson, Rodrigo and Hosang, Jan and Hein, Matthias and Schiele, Bernt},
  booktitle={2017 IEEE Conference on Computer Vision and Pattern Recognition (CVPR)}, 
  title={Simple Does It: Weakly Supervised Instance and Semantic Segmentation}, 
  year={2017},
  volume={},
  number={},
  pages={1665-1674},
  doi={10.1109/CVPR.2017.181}}

@misc{mackowiak2018cereals,
      title={CEREALS - Cost-Effective REgion-based Active Learning for Semantic Segmentation}, 
      author={Radek Mackowiak and Philip Lenz and Omair Ghori and Ferran Diego and Oliver Lange and Carsten Rother},
      year={2018},
      eprint={1810.09726},
      archivePrefix={arXiv},
      primaryClass={cs.CV},
      url={https://arxiv.org/abs/1810.09726}, 
}

@misc{arazo2020pseudo,
      title={Pseudo-Labeling and Confirmation Bias in Deep Semi-Supervised Learning}, 
      author={Eric Arazo and Diego Ortego and Paul Albert and Noel E. O'Connor and Kevin McGuinness},
      year={2020},
      eprint={1908.02983},
      archivePrefix={arXiv},
      primaryClass={cs.CV},
      url={https://arxiv.org/abs/1908.02983}, 
}

@inproceedings{vu2019advent,
  author={Vu, Tuan-Hung and Jain, Himalaya and Bucher, Maxime and Cord, Matthieu and P{\'e}rez, Patrick},
  booktitle={2019 IEEE/CVF Conference on Computer Vision and Pattern Recognition (CVPR)},
  title={ADVENT: Adversarial Entropy Minimization for Domain Adaptation in Semantic Segmentation},
  year={2019},
  pages={2512-2521},
  doi={10.1109/CVPR.2019.00262},
  url={https://doi.org/10.1109/CVPR.2019.00262}
}

@inproceedings{luo2019clan,
  author={Luo, Yawei and Zheng, Liang and Guan, Tao and Yu, Junqing and Yang, Yi},
  booktitle={2019 IEEE/CVF Conference on Computer Vision and Pattern Recognition (CVPR)},
  title={Taking a Closer Look at Domain Shift: Category-Level Adversaries for Semantics Consistent Domain Adaptation},
  year={2019},
  pages={2502-2511},
  doi={10.1109/CVPR.2019.00261},
  url={https://doi.org/10.1109/CVPR.2019.00261}
}

@misc{guan2023ilmassl,
  title={Iterative Loop Method Combining Active and Semi-supervised Learning for Domain Adaptive Semantic Segmentation},
  author={Li Guan and Xian Yuan},
  year={2023},
  eprint={2301.13361},
  archivePrefix={arXiv},
  primaryClass={cs.CV},
  url={https://arxiv.org/abs/2301.13361}
}

@inproceedings{wu2022d2ada,
  author={Wu, Tsung-Han and Liou, Yi-Syuan and Yuan, Shao-Ji and Lee, Hsin-Ying and Chen, Tung-I and Huang, Kuan-Chih and Hsu, Winston H.},
  booktitle={European Conference on Computer Vision (ECCV)},
  title={{D$^2$ADA}: Dynamic Density-aware Active Domain Adaptation for Semantic Segmentation},
  year={2022},
  pages={449--467},
  doi={10.1007/978-3-031-19818-2_26},
  eprint={2202.06484},
  archivePrefix={arXiv},
  primaryClass={cs.CV},
  url={https://doi.org/10.1007/978-3-031-19818-2_26}
}

@inproceedings{xie2022ripu,
  author={Xie, Binhui and Yuan, Longhui and Li, Shuang and Liu, Chi Harold and Cheng, Xinjing},
  booktitle={2022 IEEE/CVF Conference on Computer Vision and Pattern Recognition (CVPR)},
  title={Towards Fewer Annotations: Active Learning via Region Impurity and Prediction Uncertainty for Domain Adaptive Semantic Segmentation},
  year={2022},
  pages={8058-8068},
  doi={10.1109/CVPR52688.2022.00790},
  url={https://doi.org/10.1109/CVPR52688.2022.00790}
}

@article{ning2023madav2,
  author={Ning, Munan and Lu, Donghuan and Xie, Yujia and Chen, Dongdong and Wei, Dong and Zheng, Yefeng and Tian, Yonghong and Yan, Shuicheng and Yuan, Li},
  journal={IEEE Transactions on Pattern Analysis and Machine Intelligence},
  title={{MADAv2}: Advanced Multi-Anchor Based Active Domain Adaptation Segmentation},
  year={2023},
  volume={45},
  number={11},
  pages={13553--13566},
  doi={10.1109/TPAMI.2023.3293893},
  url={https://doi.org/10.1109/TPAMI.2023.3293893}
}

@misc{bsbaerial_zenodo,
  author       = {de Carvalho, Osmar Luiz Ferreira and de Carvalho J{\'u}nior, Osmar Ab{\'i}lio and de Albuquerque, Anesmar Olino and Guerreiro e Silva, Daniel},
  title        = {{BsB Aerial Dataset}},
  year         = {2026},
  publisher    = {Zenodo},
  doi          = {10.5281/zenodo.20635237},
  url          = {https://doi.org/10.5281/zenodo.20635237}
}

@misc{isage_zenodo,
  author       = {de Carvalho, Osmar Luiz Ferreira},
  title        = {{iSAGE}: Iterative Sparse Annotation Guided by Expert (v1.0.0)},
  year         = {2026},
  publisher    = {Zenodo},
  doi          = {10.5281/zenodo.20596185},
  url          = {https://doi.org/10.5281/zenodo.20596185}
}

@article{ren2021survey,
author = {Ren, Pengzhen and Xiao, Yun and Chang, Xiaojun and Huang, Po-Yao and Li, Zhihui and Gupta, Brij B. and Chen, Xiaojiang and Wang, Xin},
title = {A Survey of Deep Active Learning},
year = {2021},
issue_date = {December 2022},
publisher = {Association for Computing Machinery},
address = {New York, NY, USA},
volume = {54},
number = {9},
issn = {0360-0300},
url = {https://doi.org/10.1145/3472291},
doi = {10.1145/3472291},
journal = {ACM Comput. Surv.},
month = oct,
articleno = {180},
numpages = {40}
}

@article{osco2023segment,
  title={The segment anything model (sam) for remote sensing applications: From zero to one shot},
  author={Osco, Lucas Prado and Wu, Qiusheng and de Lemos, Eduardo Lopes and Gon{\c{c}}alves, Wesley Nunes and Ramos, Ana Paula Marques and Li, Jonathan and Junior, Jos{\'e} Marcato},
  journal={International Journal of Applied Earth Observation and Geoinformation},
  volume={124},
  pages={103540},
  year={2023},
  doi={10.1016/j.jag.2023.103540},
  publisher={Elsevier}
}

@article{liu2023ococ,
  title={One class one click: Quasi scene-level weakly supervised point cloud semantic segmentation with active learning},
  author={Wang, Puzuo and Yao, Wei and Shao, Jie},
  journal={ISPRS Journal of Photogrammetry and Remote Sensing},
  volume={204},
  pages={89--104},
  year={2023},
  publisher={Elsevier},
  doi={10.1016/j.isprsjprs.2023.09.002}
}

@inproceedings{kim2023adaptive,
  title={Adaptive Superpixel for Active Learning in Semantic Segmentation},
  author={Kim, Hoyoung and Oh, Minhyeon and Hwang, Sehyun and Kwak, Suha and Ok, Jungseul},
  booktitle={Proceedings of the IEEE/CVF International Conference on Computer Vision (ICCV)},
  pages={943--953},
  year={2023},
  eprint={2303.16817},
  archivePrefix={arXiv},
  primaryClass={cs.CV},
  url={https://arxiv.org/abs/2303.16817}
}

@article{li2024scribformer,
  title={{ScribFormer}: Transformer Makes CNN Work Better for Scribble-based Medical Image Segmentation},
  author={Li, Zihan and Zheng, Yuan and Shan, Dandan and Yang, Shuzhou and Li, Qingde and Wang, Beizhan and Zhang, Yuanting and Hong, Qingqi and Shen, Dinggang},
  journal={IEEE Transactions on Medical Imaging},
  volume={43},
  number={6},
  pages={2254--2265},
  year={2024},
  publisher={IEEE},
  doi={10.1109/TMI.2024.3363190}
}

@article{yang2025unimatch,
  title={{UniMatch V2}: Pushing the Limit of Semi-Supervised Semantic Segmentation},
  author={Yang, Lihe and Zhao, Zhen and Zhao, Hengshuang},
  journal={IEEE Transactions on Pattern Analysis and Machine Intelligence},
  year={2025},
  publisher={IEEE},
  doi={10.1109/TPAMI.2025.3528453}
}

@article{amershi2014power,
  title={Power to the People: The Role of Humans in Interactive Machine Learning},
  author={Amershi, Saleema and Cakmak, Maya and Knox, W. Bradley and Kulesza, Todd},
  journal={AI Magazine},
  volume={35},
  number={4},
  pages={105--120},
  year={2014},
  doi={10.1609/aimag.v35i4.2513}
}

@article{mosqueira2023humans,
  title={Human-in-the-loop machine learning: a state of the art},
  author={Mosqueira-Rey, Eduardo and Hern{\'a}ndez-Pereira, Elena and Alonso-R{\'\i}os, David and Bobes-Bascar{\'a}n, Jos{\'e} and Fern{\'a}ndez-Leal, {\'A}ngel},
  journal={Artificial Intelligence Review},
  volume={56},
  pages={3005--3054},
  year={2023},
  publisher={Springer},
  doi={10.1007/s10462-022-10246-w}
}

@article{wu2022survey,
  title={A Survey of Human-in-the-loop for Machine Learning},
  author={Wu, Xingjiao and Xiao, Luwei and Sun, Yixuan and Zhang, Junhang and Ma, Tianlong and He, Liang},
  journal={Future Generation Computer Systems},
  volume={135},
  pages={364--381},
  year={2022},
  publisher={Elsevier},
  doi={10.1016/j.future.2022.05.014}
}

@inproceedings{xu2016deep,
  title={Deep Interactive Object Selection},
  author={Xu, Ning and Price, Brian and Cohen, Scott and Yang, Jimei and Huang, Thomas S.},
  booktitle={2016 IEEE Conference on Computer Vision and Pattern Recognition (CVPR)},
  pages={373--381},
  year={2016},
  doi={10.1109/CVPR.2016.47}
}

@inproceedings{maninis2018deep,
  title={Deep Extreme Cut: From Extreme Points to Object Segmentation},
  author={Maninis, Kevis-Kokitsi and Caelles, Sergi and Pont-Tuset, Jordi and Van Gool, Luc},
  booktitle={2018 IEEE/CVF Conference on Computer Vision and Pattern Recognition (CVPR)},
  pages={616--625},
  year={2018},
  doi={10.1109/CVPR.2018.00071}
}

@inproceedings{sofiiuk2022reviving,
  title={Reviving Iterative Training with Mask Guidance for Interactive Segmentation},
  author={Sofiiuk, Konstantin and Petrov, Ilia A. and Konushin, Anton},
  booktitle={2022 IEEE International Conference on Image Processing (ICIP)},
  pages={3141--3145},
  year={2022},
  publisher={IEEE},
  doi={10.1109/ICIP46576.2022.9897365}
}

@article{huang2024medical,
  title={Segment Anything Model for Medical Images?},
  author={Huang, Yuhao and Yang, Xin and Liu, Lian and Zhou, Han and Chang, Ao and Zhou, Xinrui and Chen, Rusi and Yu, Junxuan and Chen, Jiongquan and Chen, Chaoyu and Liu, Sijing and Chi, Haozhe and Hu, Xindi and Yue, Kejuan and Li, Lei and Grau, Vicente and Fan, Deng-Ping and Dong, Fajin and Ni, Dong},
  journal={Medical Image Analysis},
  volume={92},
  pages={103061},
  year={2024},
  doi={10.1016/j.media.2023.103061}
}

@inproceedings{ke2023hqsam,
  title={Segment Anything in High Quality},
  author={Ke, Lei and Ye, Mingqiao and Danelljan, Martin and Liu, Yifan and Tai, Yu-Wing and Tang, Chi-Keung and Yu, Fisher},
  booktitle={Advances in Neural Information Processing Systems (NeurIPS)},
  year={2023},
  eprint={2306.01567},
  archivePrefix={arXiv},
  primaryClass={cs.CV},
  url={https://arxiv.org/abs/2306.01567}
}

@article{ravi2024sam2,
  title={{SAM 2}: Segment Anything in Images and Videos},
  author={Ravi, Nikhila and Gabeur, Valentin and Hu, Yuan-Ting and Hu, Ronghang and Ryali, Chaitanya and Ma, Tengyu and Khedr, Haitham and R{\"a}dle, Roman and Rolland, Chloe and Gustafson, Laura and Mintun, Eric and Pan, Junting and Alwala, Kalyan Vasudev and Carion, Nicolas and Wu, Chao-Yuan and Girshick, Ross and Doll{\'a}r, Piotr and Feichtenhofer, Christoph},
  journal={arXiv preprint arXiv:2408.00714},
  year={2024},
  eprint={2408.00714},
  archivePrefix={arXiv},
  primaryClass={cs.CV},
  url={https://arxiv.org/abs/2408.00714}
}

@inproceedings{alc2024,
  title={Active Label Correction for Semantic Segmentation with Foundation Models},
  author={Kim, Hoyoung and Hwang, Sehyun and Kwak, Suha and Ok, Jungseul},
  booktitle={Proceedings of the 41st International Conference on Machine Learning (ICML)},
  year={2024},
  eprint={2403.10820},
  archivePrefix={arXiv},
  primaryClass={cs.CV},
  url={https://arxiv.org/abs/2403.10820}
}

@article{a2lc2025,
  title={{A$^2$LC}: Active and Automated Label Correction for Semantic Segmentation},
  author={Jeon, Youjin and Cho, Kyusik and Woo, Suhan and Kim, Euntai},
  journal={arXiv preprint arXiv:2506.11599},
  year={2025},
  eprint={2506.11599},
  archivePrefix={arXiv},
  primaryClass={cs.CV},
  note={Accepted at AAAI 2026},
  url={https://arxiv.org/abs/2506.11599}
}

@article{balentacq2024,
  title={Bayesian Active Learning for Semantic Segmentation},
  author={Didari, Sima and Hu, Wenjun and Woo, Jae Oh and Hao, Heng and Moon, Hankyu and Min, Seungjai},
  journal={arXiv preprint arXiv:2408.01694},
  year={2024},
  eprint={2408.01694},
  archivePrefix={arXiv},
  primaryClass={cs.CV},
  url={https://arxiv.org/abs/2408.01694}
}

@inproceedings{csl2025,
  title={When Confidence Fails: Revisiting Pseudo-Label Selection in Semi-supervised Semantic Segmentation},
  author={Liu, Pan and Liu, Jinshi},
  booktitle={Proceedings of the IEEE/CVF International Conference on Computer Vision (ICCV)},
  year={2025},
  eprint={2509.16704},
  archivePrefix={arXiv},
  primaryClass={cs.CV},
  url={https://arxiv.org/abs/2509.16704}
}

@article{mukhoti2018,
  title={Evaluating {Bayesian} Deep Learning Methods for Semantic Segmentation},
  author={Mukhoti, Jishnu and Gal, Yarin},
  journal={arXiv preprint arXiv:1811.12709},
  year={2018},
  eprint={1811.12709},
  archivePrefix={arXiv},
  primaryClass={cs.CV},
  url={https://arxiv.org/abs/1811.12709}
}

@inproceedings{gustafsson2020,
  title={Evaluating Scalable {Bayesian} Deep Learning Methods for Robust Computer Vision},
  author={Gustafsson, Fredrik K. and Danelljan, Martin and Sch{\"o}n, Thomas B.},
  booktitle={IEEE/CVF Conference on Computer Vision and Pattern Recognition Workshops (CVPRW)},
  year={2020},
  eprint={1906.01620},
  archivePrefix={arXiv},
  primaryClass={cs.CV},
  url={https://arxiv.org/abs/1906.01620}
}

@inproceedings{aleatoric-epistemic2024,
  title={Rethinking Aleatoric and Epistemic Uncertainty},
  author={Bickford Smith, Freddie and Kossen, Jannik and Trollope, Eleanor and van der Wilk, Mark and Foster, Adam and Rainforth, Tom},
  booktitle={Proceedings of the 42nd International Conference on Machine Learning (ICML)},
  year={2025},
  eprint={2412.20892},
  archivePrefix={arXiv},
  primaryClass={cs.LG},
  url={https://arxiv.org/abs/2412.20892}
}

@article{munjal2025,
  title={Realistic Evaluation of Deep Active Learning for Image Classification and Semantic Segmentation},
  author={Mittal, Sudhanshu and Niemeijer, Joshua and {\c{C}}i{\c{c}}ek, {\"O}zg{\"u}n and Tatarchenko, Maxim and Ehrhardt, Jan and Sch{\"a}fer, J{\"o}rg P. and Handels, Heinz and Brox, Thomas},
  journal={International Journal of Computer Vision},
  volume={133},
  pages={4294--4316},
  year={2025},
  doi={10.1007/s11263-025-02372-z}
}

@misc{carvalho2026samsing,
  title={Remote {SAMsing}: From Segment Anything to Segment Everything},
  author={Carvalho, Osmar Luiz Ferreira de and Carvalho J{\'u}nior, Osmar Ab{\'i}lio de and Albuquerque, Anesmar Olino de and Silva, Daniel Guerreiro e},
  year={2026},
  eprint={2605.00256},
  archivePrefix={arXiv},
  primaryClass={cs.CV},
  url={https://arxiv.org/abs/2605.00256}
}

@inproceedings{krahenbuhl2011densecrf,
  title={Efficient Inference in Fully Connected {CRFs} with {G}aussian Edge Potentials},
  author={Kr{\"a}henb{\"u}hl, Philipp and Koltun, Vladlen},
  booktitle={Advances in Neural Information Processing Systems (NeurIPS)},
  volume={24},
  pages={109--117},
  year={2011},
  url={https://papers.nips.cc/paper/4296-efficient-inference-in-fully-connected-crfs-with-gaussian-edge-potentials}
}

\end{document}